# A Machine Learning Framework Towards Transparency in Experts' Decision Quality

**Wanxue Dong**
The Department of Information, Risk
and Operations Management
The University of Texas at Austin
Austin, TX 78705
wanxue.dong@utexas.edu

**Maytal Saar-Tsechansky**[*]
The Department of Information, Risk
and Operations Management
The University of Texas at Austin
Austin, TX 78705
maytal@mail.utexas.edu

**Tomer Geva**[*]
Coller School of Management
Tel-Aviv University
Tel Aviv-Yafo, Israel
tomergev@tauex.tau.ac.il

October 19, 2021

## Abstract

Expert workers make non-trivial decisions with significant implications. Experts' decision accuracy is thus a fundamental aspect of their judgment quality, key to both management and consumers of experts' services. Yet, in many important settings, transparency in experts' decision quality is rarely possible because ground truth data for evaluating the experts' decisions is costly and available only for a limited set of decisions. Furthermore, different experts typically handle exclusive sets of decisions, and thus prior solutions that rely on the aggregation of multiple experts' decisions for the same instance are inapplicable. We first formulate the problem of estimating experts' decision accuracy in this setting and then develop a machine-learning-based framework to address it. Our method effectively leverages both abundant historical data on workers' past decisions, and scarce decision instances with ground truth information. We conduct extensive empirical evaluations of our method's performance relative to alternatives using both semi-synthetic data based on publicly available datasets, and purposefully compiled dataset on real workers' decisions. The results show that our approach is superior to existing alternatives across diverse settings, including different data domains, experts' qualities, and the amount of ground truth data. To our knowledge, this paper is the first to posit and address the problem of estimating experts' decision accuracies from historical data with scarcely available ground truth, and it is the first to offer comprehensive results for this problem setting, establishing the performances that can be achieved across settings, as well as the state-of-the-art performance on which future work can build.

## 1 Introduction

A significant element of experts' responsibilities involves making non-trivial judgment to arrive at decisions, such as medical or engineering diagnostic decisions. Decision accuracy is a fundamental aspect of experts' judgment quality [Tetlock, 2017] and is key to both management and consumers of the experts' services. For management, assessment of expert judgment accuracies informs a slew of organizational tasks, including compensation, retention, and the efficient assignment of experts to tasks [Milkovich et al., 2011, Grote, 2005, Wang et al., 2019]. For consumers, experts' decision accuracy is key to the choice of a suitable expert and to devising risk-mitigation strategies. While poor transparency in experts' accuracy undermines both consumers' choices and the effective management of experts, assessing expert decision accuracy is challenging in these settings. This is because, in many important contexts, ground truth data — i.e., data about the correct decision for a given instance — is costly to acquire and thus rarely available. For example, physicians may determine a diagnosis and initiate a treatment, yet the correct decision — the one that can be established, such as by a panel of physicians — is most often prohibitively costly to acquire for even a small fraction of each expert's decisions. Consequently, in practice, there is often poor transparency regarding experts' decision accuracies.

---
[*]equal contribution

This research aims to develop the groundwork toward achieving scalable and inexpensive assessments of expert decision accuracies so as to promote transparency in experts' decision-making performance. We first formulate the problem and propose a novel approach for estimating experts' decision accuracies from historical data on the experts' past decisions, for which only scarce ground truth data are available. We consider settings with multiple expert workers, in which each of the workers routinely makes decisions for a mutually exclusive set of decision instances (e.g., each physician handles diagnoses for a mutually exclusive set of patients). In addition, the decision histories of each expert are known, and ground truth is available for only a scarce subset of these decisions. In principle, noisy experts' decisions can be beneficial for learning about experts' decision performances; at the same time, as more ground truth about an expert's past decisions becomes available, exclusive reliance on noisy decisions may become progressively less advantageous, relative to reliance on ground truth. In this work, we posit the problem and develop a novel method to reliably assess experts' accuracies in these settings — a method that can exploit both experts' decisions and ground truth.

In particular, we consider practical contexts in which each expert makes decisions for a mutually exclusive (non-overlapping) set of instances, and where ground truth decisions are available ex-post for a scarce subset of each of the expert's past decisions. For this setting, we consider this problem: *how can experts' decision accuracies be estimated so as to achieve better and otherwise comparable accuracy to what can be achieved by relying exclusively on ground truth decisions*. Addressing this problem calls for an approach that can be broadly applied and relied on in practice: i.e., it ought to readily apply to a broad set of data domains, and yield consistent performance, given any number of ground truth labels. To our knowledge, no prior work has considered this problem. In this paper we propose a method that effectively leverages both large amounts of historical data on the experts' noisy decisions, as well as scarce ground truth data. To yield reliable estimations under different settings and amounts of ground truth data, our approach adaptively relies more heavily on ground truth as more ground truth is available. In doing so, our method aims to offer consistently state-of-the-art estimation that can be relied on in practice.

We performed extensive empirical evaluations of our approach, comparing it to alternative approaches under a broad range of data domains and with real human expert workers. The results establish that the proposed method effectively leverages both ground truth data and noisy data, often yielding superior or at least comparable performance to alternatives across multiple data domains, different levels of workers' decision qualities, and different numbers of instances with ground truth data. We then confirm the advantageous performance of our method for real human expert workers. Our results show that our method constitutes a benchmark for future work. In particular, we find that our method yields state-of-the-art performance and offers meaningful improvements over the alternatives, while exhibiting robust performance across settings. With limited ground truth data, our approach shows up to 88% improvement over the best alternative. In addition, our approach offers consistently reliable performance across domains, across a range of experts' accuracies, and across different levels of availability of ground truth. We also conduct ablation studies to shed light on the relative contributions to performance of key elements of our approach's design.

In addition, we consider a related setting, where ground truth is not available for the decisions made by the experts being evaluated, but is available for an independent set of decision instances that were not decided by the experts being assessed. This setting corresponds to practical settings in which a panel of experts produces a small set of "exemplary" decisions, such as those made by panels of medical experts who provide carefully justified decisions for a small set of patients, often intended to inform other experts' evaluations. For these settings, we propose a variant of our approach, conduct extensive evaluations, and find that our method yields state-of-the-art performance.

To our knowledge, no prior work has considered the problem of proposing a scalable, inexpensive, and reliable means to infer costly experts' decision accuracies from historical decision data and scarce ground truth data, when workers make decisions on mutually exclusive sets of decisions. Our contributions are as follows:

1. We propose the problem of data-driven inference of experts' decision accuracies based on a large amount of historical data on experts' past decisions, where each expert makes a decision for an exclusive set of decisions and where the ground truth for these decisions is only scarsely available. To our knowledge, this is the first research to consider this problem and to propose reliable solutions for this setting.

2. We develop a novel machine-learning algorithm to address this problem. Our approach leverages both experts' past decisions and scarce instances of ground truth to yield reliable estimations across different settings. Furthermore, our approach can be readily applied to any decision data domain and can use machine-learning induction algorithms that are most suitable for learning in the corresponding domain.

3. We conducted comprehensive evaluations of our approach relative to alternatives under a wide variety of settings, including with real human workers. Our results demonstrate that our approach offers consistent, state-of-the art performance that can be relied on in practice. To our knowledge, our paper also is the first to offer comprehensive results on data-driven estimation of expert workers' decision accuracies; they demonstrate the range of performances that can be achieved across settings based on benchmark data and a new data set of human workers, and they establish the state-of-the-art performance for future work to build on.

4. In this work, we also characterize the key elements of practical settings that may affect the performance not only of our approach but also of other approaches for the problem of inferring experts' decision accuracies in our setting. The elements we identify include the underlying data domain's predictability, experts' levels of expertise, and the number of ground truth instances – all of which are key to establishing the effectiveness of data-driven solutions.

5. We propose a related problem and setting, in which ground truth data are available only for an independent set of decision instances, not considered by any of the experts being evaluated. We propose and evaluate a variant of our approach to address this setting.

This work serves as the groundwork with which to advance transparency in expert worker markets through a data-driven computational method for estimating experts' decision-making accuracies. Although an expert's decision accuracy is integral to an expert's judgment quality and expertise [Tetlock, 2017], and a meaningful assessment thereof is thus fundamental towards key managerial tasks, such as compensation or retraining, no scalable, inexpensive, and reliable means to infer experts' decision accuracies currently are available to inform managers and consumers. We hope that our work inspires future research to further advance state-of-the-art and to help promote transparency in expert markets.

## 2   Related Work

Decision accuracy is a fundamental aspect of experts' judgment quality [Tetlock, 2017] and key to informing a broad array of organizational tasks. Yet, because ground truth in such contexts is often scarce and costly to acquire, estimating experts' accuracies in a reliable and scalable manner from noisy data and limited ground truth is a challenging, open problem. To our knowledge, not prior work has addressed this problem and setting, nor offered comprehensive results on data-driven estimation of experts' decision accuracies. In this section, we discuss how different streams of prior work relate to the contributions we present here.

The most common practice for assessing worker decision quality when ground truth is scarce has been the use of peer/human evaluations [Ferris et al., 2008], such as peer or supervisor reviews or committee-based evaluations [Shanteau et al., 2002]. A large body of work over several decades has suggested and analyzed human-based approaches, such as rating relative performance and pairwise ranking [Milkovich et al., 2011]; exploring the correlations between workers' reviews to identify inconsistencies [Harris and Schaubroeck, 1988]; and examining the evidence of rating reliability and validity [Weekley and Gier, 1989]. However, such human-intensive assessments of worker decision quality, including assessments by a committee of peers, do not scale. Because we consider expert workers whose time and effort are costly, such methods are prohibitive for continuous monitoring of experts' performance over time.

Recent machine learning research has considered problems involving human and expert workers. However, most of these works considered problems and settings that meaningfully differed from those we consider here. In particular, a significant stream of work considered the problem of improving the accuracy of data labels obtained from multiple annotators, such as crowd workers [e.g., Dai et al., 2013, Dalvi et al., 2013, Rodrigues et al., 2013, Zhou et al., 2012, Wang et al., 2017]. Unlike our focus on *costly experts*, most of these works focused on inexpensive workers who perform simple, intuitive tasks and on markets characterized by inexpensive, non-expert workers [Kittur et al., 2013]. Therefore, works in this stream of research have focused on assessment methods based on repeated labeling, in which multiple workers evaluate the same data instance and the ground truth is inferred by aggregating multiple lables [e.g., Dawid and Skene, 1979, Dai et al., 2013, Dalvi et al., 2013, Kiruthiga and Sangeetha, 2016, Rodrigues et al., 2013, Sheng et al., 2017, Wang et al., 2017, Zhou et al., 2012]. Various other aspects of the repeated labeling scenario have been studied as well, including selecting instances for repeated labeling [Ipeirotis et al., 2014, Wauthier and Jordan, 2011, Zhang et al., 2014]; evaluating worker quality based on repeated labels [Raykar et al., 2010, Wang et al., 2017]; and reducing the number of labeled instances necessary to achieve a certain goal [Karger et al., 2011, 2014, Branson et al., 2017]. However, as discussed above, peer evaluations where multiple experts evaluate the same decision instances handled by other experts is not practical or feasible in our context.

Other works are related to our research because they consider predicting or assessing workers current or future performance, but these works consider settings where relevant ground truth is always available and they address different challenges than assessing experts' decision accuracies in our settings. For example, Kokkodis and Ipeirotis [2016] considered predicting future work performance based on the workers' performance history in a different domain; Kokkodis [2021] developed a method for predicting workers' skill-set-specific reputation scores in a dynamic setting; Christoforaki and Ipeirotis [2015] propose a scalable approach for technical-skill *testing* of workers, involving scalabale generation of effective test tasks/questions, based on which workers' technical skills are assessed and for which ground truth in known. Other methods considered learning predictive models from noisy (e.g., human) labels (or decisions), but did not develop methods to assess decision makers' decision accuracy with limited ground truth. For example, Brodley and Friedl [1999] and Dekel and Shamir [2009] aim to improve model learning by removing mislabeled instances. Several other works considered the cost-effective acquisition of noisy labels (typically produced by imperfect human labelers) from which to learn accurate predictive models (e.g., Huang et al. [2017], Geva et al. [2019], Gao and Saar-Tsechansky [2020]).

More recent works [Khetan et al., 2017, Tanno et al., 2019] aimed to improve model learning from noisy labels and estimate labelers' accuracies, simultaneously. While these methods did not consider how to bring to bear limited ground truth to assess workers' decision accuracy, they can apply to estimate workers' accuracies in our setting.[2] Tanno et al. [2019] showed that their approach is superior to that proposed by Khetan et al. [2017], and we thus empirically compare our approach to it. Specifically, Tanno et al. [2019] proposed minimizing the loss for models that accommodate a cross-entropy loss function, and included a regularization term based on labelers' estimated accuracies. Our approach is distinct from the method proposed by Tanno et al. [2019] by two key elements: (a) our approach is designed to leverage scarce ground truth, which the method by Tanno et al. [2019] does not take advantage of; and (b) our approach is model/domain-agnostic: it allows using the model induction algorithm most suitable for the underlying expert data domain; by contrast, Tanno et al. [2019] consider models with a cross-entropy loss function and the method is thus applicable to data domains where such models are suitable. Consequently, the worker assessment produced in Tanno et al. [2019] does not yield competitive performance in the setting we consider in this paper: we show that even for settings where our approach has the least relative advantage, with a minimal number of ground truth labels, our method yields superior assessments of experts' accuracies.

Finally, a related work on which we build [Geva and Saar-Tsechansky, 2016, 2021] proposed the problem of *ranking* expert workers according to the quality of their decisions in the absence of ground truth decisions; however, this work did not address the problem of estimating workers' decision accuracies and considered different settings than those we focus on here. In particular, ranking is a fundamentally different task than estimating workers' accuracies, and it serves different goals. While ranking aims to position workers relative to others within a cohort, an expert's absolute decision accuracy is fundamental to establish whether a given worker meets a certain performance requirement or expectation, to optimally assign workers to task, to determine whether there are practically meaningful gaps between workers' decision accuracies, and to inform retraining and compensation decisions. By contrast, the method we develop here offers novel means to reliably estimate experts' decision accuracies by leveraging noisy data along with scarce ground truth. To our knowledge, our approach produces state-of-the-art estimates unmatched by any existing alternative.

---

[2]Both works also consider the use of repeated labeling, but this scenario is not applicable in our expert setting.

## 3 Problem Formulation

We consider a set of $K$ expert workers $W = \{W_1, ..., W_K\}$, where each routinely makes multiple decisions, and where decisions made by different workers are drawn from the same distribution. For example, workers may be auditors who decide whether a given tax return claim is fraudulent or radiologists who decide whether a patient's image exhibits a certain malady. (Henceforth, we use the terms *expert workers*, *workers*, and *experts* interchangeably). We consider a challenging setting that arises often in practice, where each decision instance, such as a particular patient's diagnosis, is made by a single expert, so that the sets of decisions made by each expert are mutually exclusive. For a given expert worker, $W_k$, historical data about the worker's past decisions are available and where instance feature values instance arriving from distribution $\mathcal{X}$ is available and given by $S_{W_k} = \{X_i^k, \hat{Y}_i^k\}_{i=1}^{n_{W_k}}$. For each decision instance $i$, historical data include the worker's decision $\hat{Y}_i^k \in \{0, 1\}$, (e.g., whether or not the patient has a tumor) along with a feature vector $X_i^k \sim \mathbb{P}(\mathcal{X})$, and reflecting (possibly a subset of) the instance's feature values, such as various lab blood-test results and symptoms. For each worker $W_k$, we seek to assess the worker's decision accuracy, given by $q_{W_k} = (\sum_{i=1}^{n_{W_k}} I[Y_i^k == \hat{Y}_i^k])/n_{W_k}$, where $Y_i^k$ is the ground truth (correct) decision, and $I$ is the truth function, such that $I[\cdot] = 1$ if $(\cdot)$ is true, and $I[\cdot] = 0$, otherwise.

*Table 1: Key Notations*

| Notation | Description |
| --- | --- |
| $B_k$ | A single base model, mapping : $X^k \rightarrow \hat{Y}^k$ which is trained on worker $W_k$'s decision instances $S_{W_k}$ |
| $B_j(X_i^k)_z$ | Base model $B_j$'s probability estimate that $X_i^k$ maps to class $z$ |
| $Conf_{X_i^k}$ | The confidence in the ensemble model $M$'s prediction for instance $X_i^k$ |
| $DQ_{W_k}$ | The decision quality score for the worker $W_k$; it corresponds to the ordinal ranking of workers. |
| $f : DQ \rightarrow q$ | A learned mapping between a worker's DQ to the worker's decision accuracy |
| $GT = \bigcup_{k=1}^{K} GT_k$ | The union of decision instances with ground truth information of all workers in $W$ |
| $GT_{exc}$ | A small independent ground truth set; Instances in $GT_{exc}$ include decisions that were not made by the workers being evaluated |
| $M$ | Ensemble model $M$ |
| $n_{W_k}$ | Number of decisions made by expert worker $W_k$ |
| $q_{sw_i}$ | Decision accuracy of a synthetic worker's decision set $S_{sw_i}$ |
| $q_{W_k}$ | True decision accuracy for worker $W_k$ |
| $S_{W_k} = \{X_i^k, \hat{Y}_i^k\}_{i=1}^{n_{W_k}}$ | Worker $W_k$'s decision data |
| $S_{sw}$ | Decision data reflecting synthetic worker $sw$ |
| $t_{W_k}$ | Number of ground truth instances available decisions made by $W_k$ |
| $W = \{W_1, ..., W_k\}$ | Set of expert workers to be evaluated |

In this work we consider a challenge arising in many expert environments, where ground truth information, such as decisions produced by a panel of experts, are costly, and can thus be acquired for only a scarce subset of decisions made by each expert worker. Specifically, the set of decisions with ground truth for worker $W_k$ is given by $GT_k = \{X_i^k, Y_i^k\}_{i=1}^{t_{W_k}}$, where for all instances $X_i^k \in GT_k$: $X_i^k \sim \mathbb{P}(\mathcal{X})$, $GT_k \subseteq S_{W_k}$, and where ground truth data are scarce, - that is, $t_{W_k} \ll n_{W_k}$. Table 1 summarizes key notations used throughout the paper.

## 4 Our Approach

In this section, we outline our approach for addressing the problem above: the **M**achine-learning-based **D**ecision quality **E**stimation-**Hyb**rid (MDE-HYB) method. Our approach first produces and uses two different and complementary estimates of experts' decision accuracies, each relying on different information sets and processes that can be advantageous under different circumstances. The first estimate, the **M**achine-learning-based **D**ecision quality **E**stimation (MDE), is a machine-learning-based approach that exploits the large amount of data available on the experts' decision data, along with (scarce) ground truth information (Section 4.1, Algorithm 1). Yet, as more ground truth becomes available, an estimation that relies exclusively on ground truth decisions can yield an optimal estimation. Hence, our approach brings to bear a second estimate that is simply the frequency of correct decisions computed exclusively from ground truth data (Section 4.2), which we henceforth refer to as the **E**stimated **A**ccuracy **R**ate (EAR). The MDE-HYB method evaluates the error rates of the two estimates (MDE and EAR), and if one of the estimates is deemed superior, it selects that estimate. Otherwise, MDE-HYB infers experts' decisions accuracies as a linear combination of the two estimates (Section 4.2, Algorithm 2). Below, we outline each of the elements of our approach.

## 4.1 Machine-Learning-Driven Expert Estimation

In this section we describe the first element of our approach, MDE, which produces a machine-learning-based estimate of experts' accuracies. When ground truth is scarce, MDE aims to effectively leverage the large number of noisy decisions by expert workers, along with scarce ground truth information, to infer expert workers' decision accuracies. In principle, one can compute the rate of correct decisions for each expert worker based on the accuracy rate for the expert worker's past decisions with ground truth. However, when ground truth data are known for only a handful of each worker's decisions, the accuracy of such estimations is poor. Meanwhile, for settings in which no ground truth information is available and expert workers' true accuracies are unknown, prior work proposed a Decision Quality (DQ) score and showed that ranking decision makers by their respective DQ scores yields a ranking similar to the workers' ranking based on their true (and unknown) decision accuracy rates [Geva and Saar-Tsechansky, 2021]. However, and importantly, the DQ scores do not correspond to decision accuracy estimates - that is, they do not reflect a worker's rate of correct decisions. Thus, in our setting, computing an expert's frequency of correct decisions based on ground truth instances yields a poor estimate of the expert worker's decision accuracy, and prior work's scores yield a good ranking but do not reflect expert workers' accuracy rates; in light of these challenges, the first element of our approach, MDE, offers a computational framework that allows us to effectively leverage both the DQ scores and the limited ground truth to produce estimates of expert workers' decision accuracies (Detailed information regarding how DQ scores are computed is provided in section 4.1.2 below and in Algorithm block 1).

In particular, MDE relies on two key notions. First, workers' DQ scores can be computed without ground truth and have been shown to correlate with workers' true accuracies. Consequently, if the true accuracy, q, of some workers were somehow known, it would be possible to produce a set of $(DQ, q)$ pairs, from which it is possible to induce a mapping between a worker's DQ score and the worker's decision accuracy, $f : DQ \rightarrow q$. Such mapping could be subsequently applied to infer the decision accuracies of workers whose true decision accuracies are unknown.

The second notion that MDE builds on aims to overcome the challenge of producing the $(DQ, q)$ pairs from which a mapping between a worker's DQ score and accuracy can be learned. In particular, to induce a correct mapping, both the accuracies $q$ should be correct, and sufficient $(DQ, q)$ pairs should be available from which to reliably learn the mapping. However, in our setting, a worker's decision accuracy, $q$, is unknown; neither can it be reliably estimated based on the scarce set of instances for which ground truth is available. Recall that in the economics of our settings – in particular, the high costs of expert workers' time – the periodic acquisition of a large number of ground truth decisions for each expert worker (e.g., via a panel of costly experts, or through other costly procedures) is not feasible.

To address this challenge, we proposed an approach to exploit the available scarce ground truth to produce a large set of $(DQ, q)$ pairs, where $q$ is the true accuracy, rather than a noisy estimate. Specifically, we proposed an approach that includes three elements. (1.) Our approach first coalesces historical decision instances with ground truth from all the experts to compile a data set of ground truth instance; this data set is then used to simulate *"synthetic workers"* with predetermined decision accuracies; (2.) Our approach then produces DQ scores for the synthetic workers, and learns a mapping $f : DQ \rightarrow q$ from the $(DQ, q)$ pairs; (3.) The mapping can then apply to infer any given expert's decision accuracy from the expert's DQ score. In what follows, we discuss and outline each of these elements in turn (Section 4.1.1 - 4.1.3).

### 4.1.1 Producing decision data for synthetic workers with predetermined accuracies.

Our approach first compiles a data set of ground truth decision instances that is the union of all decision instances with ground truth from all experts: $GT = \bigcup_{k=1}^{K} GT_k$. $GT$ is then used to produce a set of semi-synthetic decision data sets, $S_{sw} = \{S_{sw_i}\}_{i=1}^{m}$, each $S_{sw_i}$ corresponds to a *synthetic worker*'s decision set and reflects a predetermined decision accuracy rate, $0.5 \leq q_{sw_i} \leq 1$. Thus, MDE produces semi-synthetic decision data, reflecting accuracy rates expected to arise in practice: Because we consider expert workers, we consider settings where experts exhibit a higher accuracy rate than can be produced by a random choice. We thus simulate semi-synthetic data sets with accuracies ranging between $[0.5, 1]$. In addition, consecutive synthetic workers' accuracies differ by a fixed (small) interval *intv*. Each set $S_{sw_i}$ contains all the instances in $GT$, and the predetermined decision accuracy $q_{sw_i}$ is produced by flipping the (correct) labels of $1 - q_{sw_i}$ proportion of instances, drawn uniformly at random from $GT$, thereby creating a $1 - q_{sw_i}$ proportion of incorrect decisions. See also lines 2-6 in Algorithm 1.[3]

### 4.1.2 Generating $(DQ, q)$ pairs and a score-accuracy mapping.

Having the synthetic workers' decision data available allows us to compute the DQ score for each decision data set $S_{sw_i}$ using the REQ method.

The REQ method [Geva and Saar-Tsechansky, 2021] computes DQ scores based on the weighted rate of agreement between the expert's decisions and the decisions inferred by an ensemble model $M$, and where each (dis)agreement is weighted by a corresponding confidence in the ensemble's prediction. Specifically, given a set of decision data sets, $S_{W_k}$, the ensemble's inferred decision, $M(X_i^k)$ for an instance $X_i^k$, is produced from the prediction of an ensemble of base models $\{B_j\}_{j=1}^{K}$, where $j \neq k$, and where each base model is a mapping of $B_j : X \to \hat{Y}$, induced from the decision data set $S_{W_j}$.

The ensemble's inferred decision is given by: $M(X_i^k) = \arg\max_z (\sum_{j=1, X_i^k \notin S_j}^{K} B_j(X_i^k)_z)$, where $B_j(X_i^k)_z$ denotes base model $B_j$'s probability estimate that $X_i^k$ belongs to the decision class $z$. Ultimately, a DQ score for a decision data set $S_k$ is given by:

$$DQ_k = \frac{\sum_{\{X_i^k, \hat{Y}_i^k\} \in S_k^+} Conf_{X_i^k}}{(\sum_{\{X_i^k, \hat{Y}_i^k\} \in S_k^+} Conf_{X_i^k}) + (\sum_{\{X_i^k, \hat{Y}_i^k\} \in S_k^-} Conf_{X_i^k})} \quad (1)$$

In Equation 1, the sets $s_k^+ \subset S_k$ and $s_k^- \subset S_k$ denote the set of a worker's decisions that agree and that disagree, respectively, with ensemble model $M$ inferred labels. (Eq. 1 could be used for either real workers or synthetic workers, so $S_k$ could be a real worker's decision set $S_{W_k}$ or a synthetic worker's decision set $S_{sw_k}$.) $Conf_{X_i^k}$ denotes ensemble $M$'s confidence in inferring the decision $X_i^k$'s, given by $Conf_{X_i^k} = \sum_{j=1, X_i^k \notin S_k}^{K} B_j(X_i^k)_{M(X_i^k)}$, where $B_j(X_i^k)_{M(X_i^k)}$ denotes $B_j$'s probability estimate that $X_i^k$ maps to the class inferred by the ensemble $M$, and $X_i^k \notin S_k$ indicates that the sum does not include estimations of a base model $B_j$ if $X_i^k$ is a member of the data set $S_k$ from which $B_j$ was induced. Thus, the confidence $Conf_{X_i^k}$ reflects a weighted count of votes of the base models towards model $M_j$'s class prediction $(X_i^k)$, where each vote is weighted by the corresponding base model's probability estimation.

Because scarce ground truth decisions are available in our settings, before we induce the base models, we replace each noisy decision $\hat{Y}_i^k$ with the corresponding ground truth decision $Y_i^k$, when it is available. In addition, in the experiments that follow, base models were produced using a random forest algorithm with 100 trees; however, base models can be induced using any classification algorithm that produces class probability estimates, and that is most suitable for the underlying experts' domain of expertise. (Lines 7-9 in Algorithm 1 MDE detail how the REQ base models are trained. Procedure "Produce DQ Score" in lines 16-22 in Algorithm 1 MDE detail how the DQ scores are calculated.)

Together, the synthetic workers' predetermined accuracies and the DQ scores, result in a data set of DQ-accuracy pairs, $\{DQ_i, q_i\}_{i=1}^{m}$, from which a mapping $f : DQ \to q$ can be learned (Lines 10-12 in Algorithm 1 MDE). In principle, any regression algorithm can be applied to learn this mapping and can be selected based on cross-validation performance. In our implementation, we use a simple linear regression, informed by the analysis in [Geva and Saar-Tsechansky, 2021], which suggests a linear relationship between a worker's DQ score and true decision accuracy rate.

---
[3]In the experiments below $N = 50$ and $intv = 0.005$

### 4.1.3 Inferring real workers' decision accuracies.

---
**Algorithm 1:** MDE
---

1 **Algorithm** MDE:
   **Input:** $\{S_{W_k}\}_1^K, GT$
   // Creating $C$ sets of $N$ synthetic workers:
2 **for** $c = 1...C$ **do**                                          /* used $C = 10$, $N = 101$ */
3    **for** $n = 1...N$ **do**
4       $q_n^c = (1 - (n-1) * intv)$       /* used $intv = 0.005$, $\{q_n^c\}_1^N$ is in range $[0.5, 1]$ */
5       $S_{sw_n}^c \leftarrow$ GT
6       $S_{sw_n}^c \leftarrow$ Randomly draw a proportion of $(1 - q_n^c)$ instances from $\mathcal{S}_{sw_n}^c$ and invert their $Y$ labels
   /* $\{\{S_{sw_n}^c\}_1^N\}_1^C$ synthetic workers created with accuracies $\{\{q_n^c\}_1^N\}_1^C$                            */
   // Training base models on real workers' data $\{S_{W_k}\}_1^K$:
7 **foreach** $S_{W_k} \in \{S_{W_k}\}_1^K$ **do**
8    **foreach** $\{X_i^k, \hat{Y}_i^k\} \in \mathcal{S}_{W_k}$ **do:** **if** $\{X_i^k, Y_i^k\} \in GT$ **then** replace $\hat{Y}_i^k$ with $Y_i^k$
9    Train base model $B(S_{W_k})$ on $S_{W_k}$
   /* K base models $\{B_j\}_1^K$ created                                                                                  */
   // Using C sets of synthetic workers to produce $C$ different mappings:
10 **for** $c = 1...C$ **do**                    /* Each mapping $c$ produced from $N$ synthetic workers */
11    $\{DQ_{sw_n}^c\}_1^N \leftarrow$ **for** $n = 1...N$ **do:** Produce DQ Scores$(S_{sw_n}^c, \{B_j\}_1^K)$
12    Induce a mapping: $f_c : DQ \rightarrow q$ from the set $\{DQ_{sw_n}^c, q_n^c\}_{n=1}^N$
   /* C mapping functions $\{f_c\}_1^C$ created                                                                           */
   // Producing DQ scores and assessments for real workers $W = \{W_1, ..., W_k\}$:
13 $\{DQ_{W_k}\}_1^K \leftarrow$ **for** $k = 1...K$ **do:** Produce DQ Score $(S_{Wk}, \{B_j\}_1^K)$
14 $\{\hat{q}_{W_k MDE}\}_1^K \leftarrow$ **for** $k = 1...K$ **do:** Produce Assessment$(S_{W_k}, \{f_c\}_1^C, DQ_{W_k})$
15 **return** $\{\hat{q}_{W_k MDE}\}_1^K, \{B_j\}_1^K, \{f_c\}_1^C$
   /* $\{\hat{q}_{W_k MDE}\}_1^K$ are Alg.1's assessments of workers' accuracies. Alg.1 is also
      applied in Alg.2, so $\{B_j\}_1^K$, $\{f_c\}_1^C$ are returned for reuse in Alg.2.                                    */

16 **Procedure** Produce DQ Score$(S_k, \{B_j\}_1^K)$:
   // Evaluating $S_k$'s DQ (real or synthetic decisions $S_{w_k}$ or $S_{sw_k}$):
17    $s_k^+ = \{\}, s_k^- = \{\}$
18    **foreach** $\{X_i^k, \hat{Y}_i^k\} \in S_k$ **do**
19       $Conf_{X_i^k} = \sum_{j=1, X_i^k \notin S_k}^K B_j(X_i^k)_{M(X_i^k)}$
20       **if** $\hat{Y}_i^k == M(X_i^k)$ **then** $s_j^+ = s_j^+ \bigcup \{X_i^k, \hat{Y}_i^k\}$
21       **else** $s_j^- = s_j^- \bigcup \{X_i^k, \hat{Y}_i^k\}$
22    **return** $DQ_k = \dfrac{\sum_{\{X_i^k, \hat{Y}_i^k\} \in S_k^+} Conf_{X_i^k}}{(\sum_{\{X_i^k, \hat{Y}_i^k\} \in S_k^+} Conf_{X_i^k}) + (\sum_{\{X_i^k, \hat{Y}_i^k\} \in S_k^-} Conf_{X_i^k})}$

23 **Procedure** Produce Assessment$(\{f_c\}_1^C, DQ_k)$:
24    **for** $c = 1...C$ **do**                    /* Apply $C$ mappings to predict a worker's accuracy */
25       $\hat{q}_c = f_c(DQ_k))$ and truncate $\hat{q}_c$ into range $[0.5, 1]$ if necessary
26    $\hat{q} = average(\{\hat{q}_c\}_1^C)$           /* Final estimate is the average of $C$ assessments */
27    **return** $\hat{q}$

---

The DQ score produced from an expert worker $W_k$'s historical data, $S_{W_k}$, is computed, and the mapping $f$ is applied to produce MDE's assessment of the (real) worker's decision accuracies. (See lines 13-14 in Algorithm 1

MDE and the procedure, "Produce Assessment," in lines 23-27 in Algorithm 1.) Finally, to reduce the variance of our estimation, the steps outlined in Sections (4.1.1.) - (4.1.3.) can be repeated using different random seeds in Section (4.1.1.). Specifically, in each repetition, the decision data of a given synthetic expert is simulated by inverting a different set of instances drawn uniformly at random. As a result of these repetitions, we produce $C$ different sets of DQ-accuracy pairs, from which $C$ different mappings, $\{f_c\}_1^C$, are learned. The final estimation of an expert's decision accuracy is then given by the average estimation produced by the $C$ mappings. Predicted quality values $\hat{q}$ are then truncated so as to range $[0.5, 1]$.

## 4.2 MDE-HYB: Balancing Estimations from Noisy Labels and from Ground Truth

The MDE method described above, is designed to leverage inferences from big, noisy decisions data, in addition to scarce ground truth data. It aims to be advantageous particularly when the scarce ground truth per expert ($\frac{|GT|}{|W|}$) cannot yield reliable estimation when it is used exclusively. However, as more ground truth data is available for each expert, an exclusive reliance on ground truth can yield optimal estimation. In particular, an alternative estimation, EAR, corresponds to estimating an expert $W_k$'s decision accuracy based on the rate of accurate decisions amongst the set $GT_k$ of decisions with ground truth; EAR is given by:

$$\hat{q}_k^{\text{EAR}} = \left(\sum_{i=1}^{|GT_k|} I[Y_i == \hat{Y}_i]\right)/|GT_k| \qquad (2)$$

In general, in different domains and given different number of ground truth instances per expert, either approach – EAR or MDE – may yield a more reliable estimation. Thus, leveraging either approach may be more appropriate in different contexts so as to produce a more reliable estimation than possible by relying exclusively on one approach, across contexts. Therefore, our proposed method, MDE-HYB, evaluates the accuracy of the estimations produced by MDE and by EAR in a given context; MDE-HYB then either selects the estimate that is superior, or infers experts' accuracies using a linear combination of both estimates. The key challenge we address is that determining the accuracy of the estimations produced by MDE and EAR in a given context is non-trivial, given workers' accuracies are unknown. Algorithm 2 MDE-HYB outlines the pseudo code for producing MDE-HYB's estimations.

Specifically, MDE-HYB first applies MDE and EAR separately to produce estimates for each worker. (Lines 2-3 in Algorithm 2.). Subsequently, to determine the accuracy of each approach's estimation, MDE-HYB generates an estimation of the distribution of errors produced by each method, MDE and EAR. If either MDE or EAR is estimated to have a statistically significant and meaningfully lower estimation error, then experts' accuracies are inferred based on the corresponding superior estimation. Otherwise, when there is no evidence that one approach is superior to another in a given context, MDE-HYB estimates an expert's accuracy as a linear combination of the estimations produced by MDE and EAR.

Because properties of the distribution of MDE's errors cannot be computed in closed form, we estimate MDE's error distributions by a form of bootstrapping. Specifically, we draw $R$ different samples from $GT$, and from each sample, we simulate decision data for $P$ additional synthetic workers, with predetermined (known) accuracies. This results in decision data for $R*P$ synthetic workers. MDE and EAR are then applied to estimate the accuracies of the synthetic workers. Comparing the assessments of MDE and EAR to the predetermined decision accuracies allows us to produce a distribution of estimation errors for each method.

Specifically, to create variations across $R$ different samples, each sample size $t$ is produced by drawing instances at random from the set of all available ground truth instances, $GT$, such that $t \leq |GT|$[4]. From each of the samples, we then simulate decision data for $P$ different synthetic workers, each with a different decision accuracy $q_{r,p}$. Each synthetic worker's accuracy $q_{r,p}$, is draw uniformly from the estimated range of the real workers' accuracies $[\hat{q}_{lower}, \hat{q}_{upper}]$. Specifically, the range $[\hat{q}_{lower}, \hat{q}_{upper}]$ reflects the 99% confidence interval of the real workers' estimated accuracies, estimated as the average estimate produced by MDE and EAR and given by $\{(\hat{q}_k^{\text{EAR}} + \hat{q}_k^{\text{MDE}})/2\}_1^K$ (Lines 14-15 in Algorithm 2). The synthetic worker's decision data is then simply generated by flipping the (correct) decisions of a $(1 - q_{r,p})$ proportion of the decision instances in the corresponding sample. Finally, for each of the $R$ samples, we apply MDE and EAR separately to produce decision accuracy estimates for $P$ synthetic workers. Because the true accuracy of each synthetic worker is known, the estimation errors for MDE and EAR can be directly computed. (The entire procedure for generating the error distributions for both MDE and EAR is detailed in lines 4, 13-30 in Algorithm 2.)

---

[4]In the experiments reported here, $t$ is either 20% of the workers' average decision data set size, or $t=|GT|$ if the former is larger than $|GT|$. Other samples sizes can be used so as to create diverse samples.

---

**Algorithm 2:** MDE-HYB

---

1 **Algorithm** `MDE-HYB`:
  **Input:** $\{S_{w_j}\}_1^K$, $GT = \{GT_k\}_1^K$
  `// Infer workers' accuracies, create base models, and mapping functions using MDE:`
2 $\{\hat{q}_k^{\text{MDE}}\}_1^K, \{B_j\}_1^K, \{f_c\}_1^C \leftarrow$ Algorithm 1: MDE($\{S_{w_j}\}_1^K$, $GT$)
  `// Infer workers' accuracies using EAR (based on EQ 2):`
3 $\{\hat{q}_k^{\text{EAR}}\}_1^K \leftarrow$ **for** $k = 1...K$ **do** : $\hat{q}_k^{\text{EAR}} = (\sum_{i=1}^{|GT_k|} I[Y_i == \hat{Y}_i])/|GT_k|$
  `// Generate distributions of MDE and EAR's errors`
4 $err_{\text{MDE}}, err_{\text{EAR}} \leftarrow$ Generate Error Dist($\{\hat{q}_k^{\text{EAR}}\}_1^K, \{\hat{q}_k^{\text{MDE}}\}_1^K, GT, \{f_c\}_1^C, \{B_j\}_1^K$))
5 $p_{EAR} \leftarrow$ Compute $p$ value: $H_0^1$:$(\mu_{\text{MDE}} - \mu_{\text{EAR}}) \leq d$  $H_a^1$:$(\mu_{\text{MDE}} - \mu_{\text{EAR}}) > d$ from $err_{\text{EAR}}, err_{\text{MDE}}$
6 $p_{MDE} \leftarrow$ Compute $p$ value: $H_0^2$:$(\mu_{\text{EAR}} - \mu_{\text{MDE}}) \leq d$  $H_a^2$:$(\mu_{\text{EAR}} - \mu_{\text{MDE}}) > d$ from $err_{\text{EAR}}, err_{\text{MDE}}$
  `// $\mu_{\text{EAR}}$ and $\mu_{\text{MDE}}$ are the distribution means of $err_{\text{EAR}}$ and $err_{\text{MDE}}$ respectively`
7 **for** $k = 1 ... K$ **do**                                             /* Estimate quality for each worker $W_j$ */
8    **if** $p_{MDE} \geq \alpha$ **and** $p_{EAR} \geq \alpha$ **then**      /* Cannot reject either hypothesis */
9      $\hat{q}_k = (\frac{p_{MDE}}{p_{MDE}+p_{EAR}} * \hat{q}_k^{\text{MDE}}) + (\frac{p_{EAR}}{p_{MDE}+p_{EAR}} * \hat{q}_k^{\text{EAR}})$
10    **else**
11      **if** $p_{MDE} < \alpha$ **then** $\hat{q}_k = \hat{q}_k^{\text{MDE}}$ **else** $\hat{q}_k = \hat{q}_k^{\text{EAR}}$
12 **return** $\{\hat{q}_k\}_1^K$

13 **Procedure** `Generate Error Dist`($\{\hat{q}_k^{\text{EAR}}\}_1^K, \{\hat{q}_k^{\text{MDE}}\}_1^K, GT, \{f_c\}_1^C, \{B_j\}_1^K$):
14   $\{\hat{q}_{k_{\text{AVG}}}\}_1^K = \frac{1}{2}\{\hat{q}_k^{\text{EAR}} + \hat{q}_k^{\text{MDE}}\}_1^K$
15   $[\hat{q}_{lower}, \hat{q}_{upper}] \leftarrow \hat{\mu} \pm t_{\alpha=0.01, K-1} \frac{\hat{\theta}}{\sqrt{n}}$    /* $\hat{\mu}$ is mean and $\hat{\theta}$ is std of $\{\hat{q}_{k_{\text{AVG}}}\}_1^K$ */
16   $err_{\text{EAR}} = \{\}$; $err_{\text{MDE}} = \{\}$
17   **for** $r = 1 ... R$ **do**                                          /* Draw $R$ different subsets of $GT$ */
18    $S_{sw_r} \leftarrow$ Randomly draw $t$ instances from $GT$
19    **for** $p = 1 ... P$ **do**                                    /* Repeat for $P$ synthetic workers */
20      $q \leftarrow$ Uniformly draw from $[\hat{q}_{lower}, \hat{q}_{upper}]$
       `// Create synthetic workers' data, apply MDE, and estimate its errors:`
21      $S_{sw_{\text{MDE}}} \leftarrow S_{sw_r}$
22      Flip each label in $S_{sw_{\text{MDE}}}$ with probability $q$
23      $DQ \leftarrow$ Produce DQ Score ($S_{sw_{\text{MDE}}}, \{B_j\}_1^K$)                 /* Alg.1 Procedure */
24      $\hat{q}_{\text{MDE}} =$ Produce Assessment ($S_{sw_{\text{MDE}}}, \{f_c\}_1^C, DQ$)      /* Alg.1 Procedure */
25      $err_{\text{MDE}} \leftarrow err_{\text{MDE}} \bigcup \{|q - \hat{q}_{\text{MDE}}|\}$
       `// Simulate decision errors, apply EAR and estimate its errors:`
26      $GT_{PE} = \frac{|GT|}{|W|}$              /* average number of ground truth per expert */
27      $S_{sw_{EAR}} \leftarrow$ Randomly draw $GT_{PE}$ instances from $S_{sw_r}$ and flip each of their labels with probability $q$
28      $\hat{q}_{\text{EAR}} \leftarrow$ Apply EAR to $S_{sw_{EAR}}$
29      $err_{\text{EAR}} \leftarrow err_{\text{EAR}} \bigcup \{|q - \hat{q}_{\text{EAR}}|\}$
30   **return** $err_{\text{MDE}}, err_{\text{EAR}}$

---

We now have the error distribution for both EAR and MDE so as to assess whether one approach yields a superior error than the other. Specifically, we examine whether the difference in errors is greater than $d$, where $d$ reflects a meaningful difference in the relevant context in practice.[5] We do so via two 2-sample one-tailed t-tests, comparing the error means of the two approaches. Specifically, in the first test, the null hypothesis is given by $H_0^1 : (\mu_{\text{MDE}} - \mu_{\text{EAR}}) \leq d$, with an alternative hypothesis of $H_a^1 : (\mu_{\text{MDE}} - \mu_{\text{EAR}}) > d$; in the second test, the null hypothesis is $H_0^2 : (\mu_{\text{EAR}} - \mu_{\text{MDE}}) \leq d$, with an alternative hypothesis of $H_a^2 : (\mu_{\text{EAR}} - \mu_{\text{MDE}}) > d^6$. (See lines 5-6 in Algorithm 2.)

---

[5]In the empirical results we report, $d = 0.01$; for example, a 1% difference in diagnosis accuracy has significant consequences in rare disease diagnosis.

Finally, if one of the two (but not both) null hypotheses is not supported, MDE-HYB uses the superior approach to infer experts' decision accuracies. Alternatively, if both null hypotheses cannot be rejected, and if the two error means are comparable in the relevant context, MDE-HYB estimates an expert's accuracy as a linear combination of MDE's and EAR's estimations. In particular, the larger $p$-value weights the estimation hypothesized to have a smaller error. (See lines 7-12 in Algorithm 2.)

## 5 Empirical Evaluations

To evaluate our approach, we conducted extensive empirical evaluations using simulation studies based on benchmark data sets and evaluations using real human decision makers. The simulation studies offer controlled settings to establish the relative performance of our approach compared to alternatives and to assess the contributions of key elements of our approach to its performance. We complement these results with evaluations of real human workers' decision accuracies. In the following sections, we outline the key elements of our empirical evaluations and discuss our results.

### 5.1 Evaluations with semi-synthetic data sets

We follow prior work and use existing data sets to simulate workers' decision data [see, e.g., Raykar et al., 2010, Ipeirotis et al., 2014, Geva and Saar-Tsechansky, 2021]. We then apply alternative methods to evaluate the degree to which they are able to correctly recover the workers' decision accuracies. Evaluations with such semi-synthetic data are necessary to establish the methods' performances under a variety of controlled settings. Specifically, because MDE-HYB relies on induced mappings from the data, our evaluations first aim to establish whether MDE-HYB is advantageous and offers robust performance across data domains, including in particular data sets that have different levels of predictability. Because MDE-HYB takes advantage of ground truth data, the controlled studies allow us to explore whether our approach can be relied on in practice across settings involving different levels of availability of ground truth. The evaluations we report also aim to explore whether MDE-HYB yields a state-of-the-art assessment of experts' decision accuracies when the experts exhibit particular properties, including, specifically, when experts exhibit different levels of decision qualities and when they exhibit correlated decision errors. In addition to evaluating our approach's performance relative to alternatives, we use these settings to conduct ablation studies, assessing the contributions of key elements of our approach to its performance. Below, we discuss the complete details of the simulation studies using semi-synthetic data sets.

We use three data sets for our empirical evaluations with semi-synthetic data sets. Our primary dataset consists of sales tax audit data [Geva and Saar-Tsechansky, 2021, Saar-Tsechansky and Provost, 2004] comprising approximately 30,000 instances and reflecting information on firms and whether they underreported their sales for tax purposes. The tax audit dataset is challenging because models induced from it to infer a firm's compliance yield relatively poor predictive performance, with an area under the ROC of about 0.67. This poor performance occurs because of the limited set of features avaiable, compared to the information needed to determine compliance in tax audits. The covariates include information such as the age of the firm, the industry, employee wages relative to peer companies, and prior audit history.

In addition, we use two publicly available datasets. The first is the Kaggle IMDb Movie review dataset with 25,000 reviews, where the independent variable reflects a human assessment of whether or not the review is positive. This data set also offers higher predictability than the tax audit data, with an area under the ROC of 0.806. In the experiments reported here, we use a simple, binary, bag-of-words representation of the reviews.[7] The third data domain included in our evaluations is the UCI Spam dataset [Bache and Lichman, 2013], which has been used in related prior works involving workers' decisions [e.g., Ipeirotis et al., 2014, Geva and Saar-Tsechansky, 2021]. This domain offers high predictability, with an area under the ROC of 0.987. As we previously noted, evaluations with datasets that have different properties and predictability aim to assess the robustness of our method's performance across data domains.

---

[6]Note that we use the two tests above because in either case the alternative hypothesis states that the mean error either of MDE or of EAR is meaningfully larger than that of the other, but not the other way around. In addition, in the empirical results reported below, $\alpha = 0.02$.

[7]We produced a binary bag-of-words representation of the dataset, reflecting 2,000 of the most often-occuring words. In general, as our results reflect, a richer representation is likely to yield a better performance of our approach, given that our method benefits from accurate mapping of the domain.

We also followed prior work by producing different levels of experts' accuracies to assess the robustness of our approach to this aspect of the setting [see, e.g., Raykar et al., 2010, Ipeirotis et al., 2014, Geva and Saar-Tsechansky, 2021]. Specifically, we evaluated each approach's ability to infer 20 workers' accuracies when workers exhibit both a *high* level of accuracies, ranging between 76% and 95%, and a lower level of accuracies, ranging between 61% and 80%, with a minor, 1% accuracy difference between consecutive workers' accuracies. To do so, we randomly partitioned each data set into $K$ exclusive, equal-sized subsets; each was then used to simulate decisions made by a different expert worker. To produce decisions for a worker with predetermined decision accuracy $g$, the correct label of each decision instance was flipped with probably $1 - g$.[8]

We also aimed to assess whether our approach can be relied on in practice to yield superior, or at least comparable, performance to the alternatives, regardless of the number of instances with ground truth. For each of the settings, we thus evaluate the performance of our method and each of the alternatives when ground truth is available for a different number of instances. Across different experiments, for each expert and for a given number of instances with available ground truth, the instances for which ground truth is known were drawn uniformly at random.

Finally, we report the average mean absolute error (MAE) between a given approach's assessment and the true accuracies of the workers. In each evaluation, we report the average MAE for each approach over 50 random experiments, along with the statistical significance of the difference in mean performance between our approach and each of the alternatives.

As we noted above, evaluations with semi-synthetic datasets have been used in multiple prior works because they are essential in establishing the robustness of a methodology under a wide variety of settings (e.g., workers' qualities, availability of ground truth, diverse data domains, etc.) and in establishing the degree to which such different conditions affect the method's relative performance relative to alternatives. These simulation studies are also necessary to assess our approach's performance when workers exhibit particular properties, such as when workers' decisions are correlated. Nevertheless, any simulation-based evaluations introduce the risk that the results may be affected by idiosyncrasies of the simulation procedure. We therefore also conducted an additional evaluation procedure on a purposely compiled human workers' decision dataset, which we compiled via Amazon Mechanical Turk. These evaluations are discussed next.

## 5.2 Evaluation on a Purposely Compiled Human Workers' Decision Dataset

In addition to the evaluations above, we evaluate each approach's performance when assessing real human workers' decision accuracies.

We recruited 40 crowd workers through Amazon Mechanical Turk (AMT). Each worker was assigned an exclusive set of 200 online reviews, and they were asked to determine whether each of the reviews assigned to them conveys a positive view or negative view of the corresponding product. The reviews were drawn from the "Amazon fine food reviews" dataset that is available on the Kaggle website,[9] for which the correct review sentiments are provided. This information allowed us to assess the AMT workers' accuracies. In Appendix D, we provide extensive details on how this data set was compiled and statistics on workers' (true) accuracy distribution.

As before, for the experiments with human workers, we also evaluated each approach's performance for settings in which it has access to a different number of ground truth instances for the reviews provided to each worker. For each such setting, we repeated the evaluation 50 times; in each repetition, ground truth was available for a different set of randomly drawn reviews assessed by each worker. The advantage of using this evaluation, versus a simulation approach with semi-synthetic datasets, is that it is not affected by the idiosyncrasies of the simulation procedure. However, the purposely compiled dataset naturally has some limitations as well, including a single, specific, distribution of workers' accuracies and a single assignment of decision instances to workers. Furthermore, although in practice experts' historical data likely include many past decisions, the costs of acquiring workers' input via AMT mean that these data include a fairly small number of decision instances per worker (i.e., 200), which in turn makes inducing productive machine learning models more difficult.

---

[8]Recall that, in all the experiments reported here, the methods being evaluated have access to only a limited set of ground truth labels in each experiment, and the workers' noisy labels are used for all other cases.

[9]https://www.kaggle.com/snap/amazon-fine-food-reviews

### 5.3 Benchmarks

We evaluated our proposed approach relative to three benchmarks that bring to bear ground truth data to estimate workers accuracies. The first benchmark is perhaps the most natural one, where an expert's accuracy rate is considered a binomial variable, and its maximum likelihood estimate is simply the rate of correct decisions among the subset of this worker's instances that have ground truth labels. Note that this natural benchmark is equivalent to the Estimated Accuracy Rate (EAR) measure, defined in Equation 2, which is also a component in the MDE-HYB approach. Henceforth, we refer to this benchmark as EAR as well.

Two additional benchmarks make use of a predictive model to evaluate workers' accuracies, as does MDE-HYB. Specifically, an expert's accuracy is estimated using the rate of agreement between the expert's decisions and those inferred by a predictive model. In the first alternative, a single "global" predictive model is induced exclusively from the set $GT$ set of ground truth instances, and the global model is then used to infer the likely correct labels to which expert decisions are compared. Henceforth, we refer to this benchmark as Global-Model-Ground-Truth (GM-GT). The second benchmark is one where an expert's decisions are compared to those inferred by a global model induced from the set $S$ of historical decision data of all experts (regardless of the availability of GT). To improve this model's induction, ground truth decisions are used instead of the expert's noisy decisions when they are available. Henceforth, we refer to this variant as Global-Model-All (GM-ALL). With both the GM-GT and GM-ALL benchmarks, an expert's accuracy is computed as the rate of agreement between the expert's decision and either the ground truth decision, when it is available for a given instance, or the model's predicted decision.

## 6 Results

In this section, we report the results of empirical evaluations, comparing our approach's performances relative to each alternative under different settings; we also report the results of ablation studies that evaluate the relative contributions of key elements of our proposed method. Consequently, we assess and report MDE-HYB's performance: (a) for different data domains, (b) when ground truth is available for different numbers of decision instances, and (c) when workers exhibit different levels of expertise.

Figure 1 (top row: low-quality workers) shows curves of the average MAE achieved by MDE-HYB, along with that achieved by the benchmarks (EAR, GM-GT, and GM-ALL), for settings with scarce truth instances, and when workers' decision accuracies range between 61% and 80% (henceforth referred to as low-quality workers). Table 2 shows the result of these experiments, along with the improvement achieved by MDE-HYB relative to each of the benchmarks and its statistical significance.[10] Importantly, while our focuses are settings with scarce ground truth instances, the tables present results with both scarce and abundant ground truth instances to determine whether MDE-HYB may be inferior and thus undesirable when ground truth is in abundance. Finally, Figure 1 bottom row and Table 2 show these results for settings with higher quality workers, whose decision accuracies range between 76% and 95%.

As shown in Figure 1 (top row: low-quality workers) and Table 2, when ground truth data are scarce, MDE-HYB achieves significantly superior estimations of workers' decision accuracies as compared to each of the alternatives, across domains and levels of workers' expertise, and it is never the worst approach for inferring experts' accuracies. For example, assuming five ground truth instances per worker, MDE-HYB achieves between 61.9% and 93.8% higher accuracy relative to the alternatives across the three domains. For the audit data set, where all methods produced the highest estimation errors, the best alternative, EAR, exhibits an average 14.2% error; the worst alternatives, GM-ALL, yields a 29.5% error; and MDE-HYB exhibits an average error of only 4.1%.

---

[10]Note that because of space constraints, results reported in tables throughout this paper are shown with only two decimal points; as a result, in a few cases, the reported difference between two methods is slightly different than if the two respective (truncated) numbers in the table are subtracted.

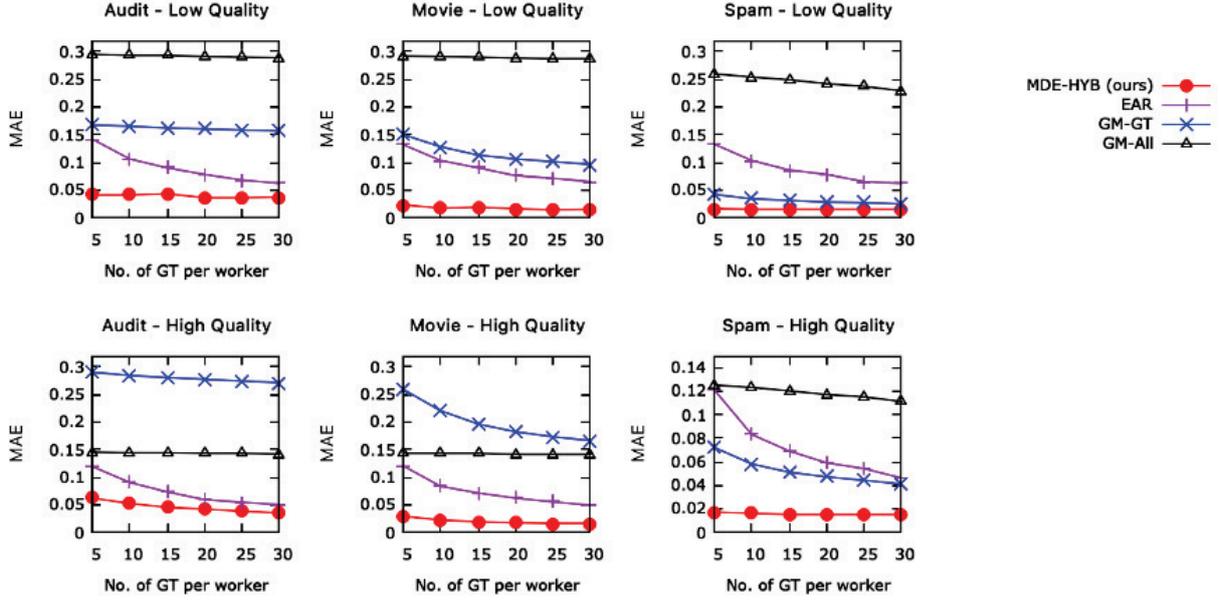

*Figure 1: MDE-HYB's Performance Relative to Benchmarks*

Table 2: MDE-HYB and Benchmarks Performance measured by MAE for Low-quality Workers

| DATASET | GT PER WORKER | MDE-HYB (ours) | EAR | MDE-HYB IMPROV | GM-GT | MDE-HYB IMPROV | GM-ALL | MDE-HYB IMPROV |
|---|---|---|---|---|---|---|---|---|
| Audit | 5 | **0.041** | 0.142 | 71.1%** | 0.167 | 75.4%** | 0.295 | 86.1%** |
| | 10 | **0.041** | 0.106 | 61.3%** | 0.165 | 75.2%** | 0.293 | 86.0%** |
| | 15 | **0.043** | 0.09 | 52.2%** | 0.162 | 73.5%** | 0.293 | 85.3%** |
| | 20 | **0.036** | 0.078 | 53.8%** | 0.16 | 77.5%** | 0.291 | 87.6%** |
| | 25 | **0.036** | 0.068 | 47.1%** | 0.158 | 77.2%** | 0.29 | 87.6%** |
| | 30 | **0.037** | 0.062 | 40.3%** | 0.157 | 76.4%** | 0.289 | 87.2%** |
| | 50 | **0.043** | 0.047 | 8.5%** | 0.153 | 71.9%** | 0.285 | 84.9%** |
| | 100 | **0.034** | 0.034 | 0.0% | 0.145 | 76.6%** | 0.275 | 87.6%** |
| | 300 | **0.019** | 0.019 | 0.0% | 0.121 | 84.3%** | 0.235 | 91.9%** |
| Movie | 5 | **0.023** | 0.132 | 82.6%** | 0.15 | 84.7%** | 0.292 | 92.1%** |
| | 10 | **0.017** | 0.103 | 83.5%** | 0.128 | 86.7%** | 0.291 | 94.2%** |
| | 15 | **0.019** | 0.09 | 78.9%** | 0.113 | 83.2%** | 0.29 | 93.4%** |
| | 20 | **0.016** | 0.076 | 78.9%** | 0.106 | 84.9%** | 0.288 | 94.4%** |
| | 25 | **0.014** | 0.071 | 80.3%** | 0.101 | 86.1%** | 0.287 | 95.1%** |
| | 30 | **0.015** | 0.065 | 76.9%** | 0.096 | 84.4%** | 0.287 | 94.8%** |
| | 50 | **0.013** | 0.05 | 74.0%** | 0.085 | 84.7%** | 0.282 | 95.4%** |
| | 100 | **0.015** | 0.035 | 57.1%** | 0.074 | 79.7%** | 0.27 | 94.4%** |
| | 300 | **0.011** | 0.018 | 38.9%** | 0.054 | 79.6%** | 0.224 | 95.1%** |
| Spam | 5 | **0.016** | 0.133 | 88.0%** | 0.042 | 61.9%** | 0.26 | 93.8%** |
| | 10 | **0.015** | 0.103 | 85.4%** | 0.034 | 55.9%** | 0.254 | 94.1%** |
| | 15 | **0.015** | 0.086 | 82.6%** | 0.031 | 51.6%** | 0.249 | 94.0%** |
| | 20 | **0.015** | 0.078 | 80.8%** | 0.028 | 46.4%** | 0.242 | 93.8%** |
| | 25 | **0.015** | 0.065 | 76.9%** | 0.027 | 44.4%** | 0.237 | 93.7%** |
| | 30 | **0.015** | 0.062 | 75.8%** | 0.025 | 40.0%** | 0.229 | 93.4%** |
| | 50 | **0.014** | 0.044 | 68.2%** | 0.022 | 36.4%** | 0.208 | 93.3%** |
| | 100 | **0.014** | 0.027 | 48.1%** | 0.014 | 0.0% | 0.15 | 90.7%** |

Experts' accuracy estimation errors. Values show Mean Absolute Error (MAE). MDE-HYB IMPROV shows the improvement of MDE-HYB over the alternative; MDE-HYB yields substantially better and otherwise comparable estimations of experts' accuracies. ** MDE-HYB is statistically significantly better ($p < 0.05$), *: ($p < 0.1$).

Note also that, given MDE-HYB's use of inference, its performance relates also to the predictability of a given domain. It exhibits an error between 4.1% and 1.9% for the audit domain, which has low predictability (AUC of 0.671), and an error between 1.6% and 1.4% for the spam domain, for which the AUC is 0.987. Interestingly, as with MDE-HYB, all benchmarks take advantage of ground truth data; that is, both GM-GT and GM-ALL include the *GT* set to induce the global model, and GM-ALL also makes use of noisy experts' decisions. Yet, MDE-HYB more effectively exploits inference, ground truth data, and experts' noisy decisions, resulting in a consistently advantageous performance across settings that is unmatched by any of the alternatives. Our results also underscore a key aspect of our approach's performance: MDE-HYB's robustness across settings involving different levels of availability of ground truth. Note that, given a sufficiently large number of instances of ground truth information, EAR is guaranteed to converge to the correct decision accuracy of a given worker. However, as shown in Figure 1 and Tables 2 and 3, all the methods' estimations improve with more ground truth, but MDE-HYB consistently either exhibits significantly superior accuracies or is otherwise comparable to the best alternative. Importantly, recall that MDE-HYB aims to estimate workers' decision performances under scarce ground truth, yet it can be safely deployed to yield state-of-the-art performance, regardless of the number of available ground truth instances. In fact, when there is abundant ground truth, both MDE-HYB and EAR, in particular, achieve comparable and highly accurate estimations.

Overall, MDE-HYB exhibits robust performance, consistently producing either the best, or at least comparable, estimations of experts' decision accuracies relative to the alternatives; these results hold across domains, across the number of ground truth instances, and across the workers' level of expertise. Figure 1 (bottom row: high-quality workers) and Table 3 show MDE-HYB's robust performance in settings with high-quality workers. As shown, when ground truth are scarce, no alternative achieves a performance comparable to that of MDE-HYB. Across all domains and levels of availability of ground truth, MDE-HYB is either superior to the best alternative or otherwise comparable.

Table 3: MDE-HYB and Benchmarks Performance measured by MAE for High-quality Workers

| DATASET | GT PER WORKER | MDE-HYB (ours) | EAR | MDE-HYB IMPROV | GM-GT | MDE-HYB IMPROV | GM-ALL | MDE-HYB IMPROV |
|---|---|---|---|---|---|---|---|---|
| Audit | 5 | **0.062** | 0.119 | 47.9%** | 0.29 | 78.6%** | 0.145 | 57.2%** |
|  | 10 | **0.052** | 0.09 | 42.2%** | 0.284 | 81.7%** | 0.144 | 63.9%** |
|  | 15 | **0.045** | 0.073 | 38.4%** | 0.28 | 83.9%** | 0.144 | 68.8%** |
|  | 20 | **0.042** | 0.059 | 28.8%** | 0.277 | 84.8%** | 0.143 | 70.6%** |
|  | 25 | **0.038** | 0.054 | 29.6%** | 0.274 | 86.1%** | 0.143 | 73.4%** |
|  | 30 | **0.035** | 0.05 | 30.0%** | 0.271 | 87.1%** | 0.142 | 75.4%** |
|  | 50 | **0.036** | 0.037 | 2.7% | 0.264 | 86.4%** | 0.14 | 74.3%** |
|  | 100 | **0.026** | 0.026 | 0.0% | 0.252 | 89.7%** | 0.135 | 80.7%** |
|  | 300 | **0.014** | 0.014 | 0.0% | 0.211 | 93.4%** | 0.116 | 87.9%** |
| Movie | 5 | **0.029** | 0.12 | 75.8%** | 0.259 | 88.8%** | 0.143 | 79.7%** |
|  | 10 | **0.022** | 0.083 | 73.5%** | 0.22 | 90.0%** | 0.143 | 84.6%** |
|  | 15 | **0.019** | 0.071 | 73.2%** | 0.196 | 90.3%** | 0.143 | 86.7%** |
|  | 20 | **0.017** | 0.062 | 72.6%** | 0.182 | 90.7%** | 0.141 | 87.9%** |
|  | 25 | **0.016** | 0.055 | 70.9%** | 0.173 | 90.8%** | 0.141 | 88.7%** |
|  | 30 | **0.016** | 0.049 | 67.3%** | 0.166 | 90.4%** | 0.141 | 88.7%** |
|  | 50 | **0.014** | 0.039 | 64.1%** | 0.147 | 90.5%** | 0.138 | 89.9%** |
|  | 100 | **0.015** | 0.026 | 42.3%** | 0.127 | 88.2%** | 0.132 | 88.6%** |
|  | 300 | **0.009** | 0.014 | 35.7%** | 0.095 | 90.5%** | 0.109 | 91.7%** |
| Spam | 5 | **0.017** | 0.121 | 86.0%** | 0.072 | 76.4%** | 0.125 | 86.4%** |
|  | 10 | **0.016** | 0.083 | 80.7%** | 0.058 | 72.4%** | 0.123 | 87.0%** |
|  | 15 | **0.015** | 0.069 | 78.3%** | 0.051 | 70.6%** | 0.12 | 87.5%** |
|  | 20 | **0.015** | 0.059 | 74.6%** | 0.047 | 68.1%** | 0.117 | 87.2%** |
|  | 25 | **0.015** | 0.054 | 72.2%** | 0.044 | 65.9%** | 0.115 | 87.0%** |
|  | 30 | **0.015** | 0.046 | 67.4%** | 0.041 | 63.4%** | 0.111 | 86.5%** |
|  | 50 | **0.015** | 0.036 | 58.3%** | 0.034 | 55.9%** | 0.101 | 85.1%** |
|  | 100 | **0.012** | 0.02 | 40.0%** | 0.022 | 45.5%** | 0.072 | 83.3%** |

Experts' accuracy estimation errors. Values show Mean Absolute Error (MAE). MDE-HYB IMPROV shows the improvement of MDE-HYB over the alternative; MDE-HYB yields substantially better and otherwise comparable estimations of experts' accuracies. ** MDE-HYB is statistically significantly better ($p < 0.05$), *: ($p < 0.1$).

## 6.1 Evaluation on Purposely Compiled Human Workers Decision Dataset

We applied MDE-HYB to evaluate the decisions accuracy of human workers recruited via Amazon Mechanical Turk (AMT) to determine the sentiments expressed in a product reviews. Figure 2 and Table 4 show performance comparisons of MDE-HYB and the benchmarks. Recall that, because of the cost of acquiring workers' decisions, these data likely include a smaller number of decision instances for each worker than is available from workers' histories in many settings in practice. As a result, this factor may undermine the effectiveness of machine learning models induced from the data. Nevertheless, as we show below, these results establish the robustness of our approach and corroborate the conclusions drawn from the results reported previously. In particular, the results establish that MDE-HYB yields state-of-the-art performance, yielding consistently and statistically significant better estimations than the alternatives, or otherwise estimations that are comparable to any of the existing alternatives.

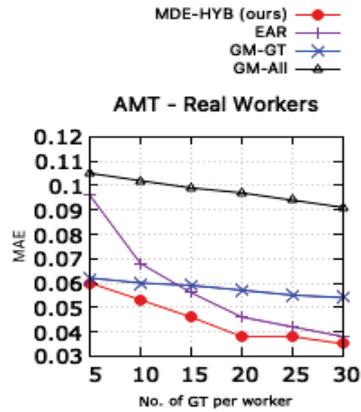

*Figure 2: Performance with AMT Real Workers*

*Table 4: MDE-HYB and Benchmarks Performance measured by MAE with Real Workers*

| GT PER WORKER | MDE-HYB (ours) | EAR | MDE-HYB IMPROV | GM-GT | MDE-HYB IMPROV | GM-ALL | MDE-HYB IMPROV |
|---|---|---|---|---|---|---|---|
| 5 | **0.060** | 0.096 | 37.2%** | 0.062 | 3.60% | 0.105 | 42.7%** |
| 10 | **0.053** | 0.068 | 22.4%* | 0.060 | 12.0% | 0.102 | 47.9%** |
| 15 | **0.046** | 0.056 | 17.9%* | 0.059 | 22.4%* | 0.099 | 54.1%** |
| 20 | **0.038** | 0.046 | 18.1%* | 0.057 | 33.3%** | 0.097 | 60.7%** |
| 25 | **0.038** | 0.042 | 9.1% | 0.055 | 31.8%** | 0.094 | 59.9%** |
| 30 | **0.035** | 0.038 | 7.1% | 0.054 | 34.5%** | 0.091 | 61.5%** |
| 50 | **0.027** | 0.027 | -0.2% | 0.047 | 41.4%** | 0.081 | 66.2%** |
| 100 | **0.015** | 0.015 | 0.0% | 0.031 | 50.6%** | 0.054 | 71.9%** |

Experts' accuracy estimation errors. Values show Mean Absolute Error (MAE). MDE-HYB IMPROV shows the improvement of MDE-HYB over the alternative; MDE-HYB yields substantially better and otherwise comparable estimations of experts' accuracies. ** MDE-HYB is statistically significantly better ($p < 0.05$), *: ($p < 0.1$).

## 6.2 Ablation Studies

The empirical evaluations demonstrate that MDE-HYB takes advantage of data-driven inference from different experts, as well as the calibration of experts' accuracies, and it relies both on experts' noisy decisions and ground truth in a manner that is unmatched by the use of this information or by benchmark methods. In the ablation studies that follow, we aim to establish the relative benefits of key elements of our approach. Specifically, we intend to establish the benefits from learning and aggregating the inference from each individual expert's and, separately, the benefits of learning both from expert's noisy decisions and from ground truth data.

We first explore whether MDE-HYB's inference of $M(X_i^j)$ is beneficial. In particular, we explore the benefit that stems from MDE-HYB's inference's being based on an ensemble of base models, each induced from a different expert's data; this appraoch allows us to account for the idiosyncratic uncertainties of different base models while aggregating their inferences.

As an alternative, we consider a variant of our approach, which follows Algorithms 1 and 2, except that in this variant, $M(X_i^k)$ is inferred from a single global model, induced from $\{S_{W_k}\}_1^K$. That is, it is inferred from all experts' noisy decision instances. We refer to this variant as MDE-GlobalModel-All, or MDE-GM-ALL. A second variant aims to explore whether greater benefit can be achieved by inferring $M(X_i^k)$ using a model induced exclusively from instances with GT, thereby avoiding learning from the experts' noisy decisions altogether. Given the scarcity of GT in our setting, including base models from only a handful of ground truth instances (e.g., 5) would not be feasible; hence, for this variant we also replace the ensemble with a single model, induced only from $GT = \bigcup_{k=1}^{K} GT_k$. We refer to this variant as MDE-SingleModel-GroundTruth, or MDE-SM-GT. Finally, for both variants, given $Conf_{X_i^k}$ can no longer be the summation of the base models' probabilities, $Conf_{X_i^k}$ is simply the global model's estimated probability for the predicted class.

For settings with scarce ground truth, Figure 3 shows the MDE-HYB's performance relative to that of each of the above variants, for experts of either low quality (top row), higher quality (bottom row), and AMT workers (bottom row, rightmost plot). Tables A1 to A3 in Appendix A provides detailed numerical results and statistical significance tests for these settings, as well as for a setting with abundant ground truth data.

The results show that inferring $M(X_i^k)$ from a global model that is either induced exclusively from $GT$ (MDE-SM-GT) or from both $GT$ and $S$ (MDE-GM-ALL) almost always yields substantially and statistically significant worse performance than that exhibited by MDE-HYB.

Of particular interest is MDE-HYB's use of noisy labels. Of the two variants of our approach, MDE-GM-ALL – which uses all of the experts' (noisy) decisions – often yields the worst performance and also can yields worse results than when relying exclusively on ground truth exclusively. In contrast, MDE-HYB's use of noisy decisions yields a superior performance relative to both variants.

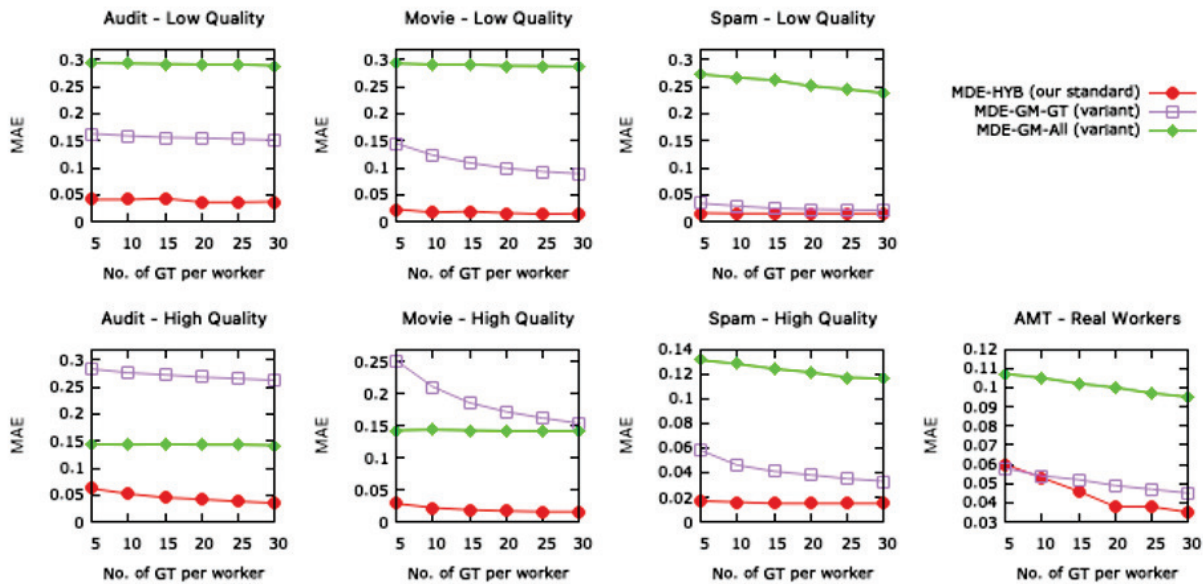

*Figure 3: Evaluating Variants of MDE-HYB*

These results reveal how MDE-HYB's learning of individual experts' decision patterns, which allows to account for uncertainties in each expert's base model inferences (reflected in $conf_{X_i^k}$), enables MDE-HYB to more effectively exploit experts' noisy decisions than is possible with a model that is simply induced from experts' noisy decisions. In addition, the variant of our approach that infers $M(X_i^k)$ using a global model, induced exclusively from the set of ground truth instances $GT$, similarly does not match MDE-HYB's performance; this underscore that MDE-HYB's particular use of experts' noisy decisions renders them instrumental for producing better performance than is possible when learning comes exclusively from instances with ground truth.

### 6.3 Additional Evaluations

#### 6.3.1 Comparisons to a model-specific alternative with no ground truth.

As discussed in the Related Work section, Tanno et al. [2019] develop a method that is designed to address a different problem and in different settings from the ones we consider here, but that can be applied to infer workers' qualities. As we discussed above, the method has several limitations which render its performance in our problem settings noncompetitive, including in particular that it does not exploit the availability of limited ground truth. Hence, in Appendix C, we provide a comparison between this approach and MDE-HYB in settings that are least advantageous to our approach. Our results, reported in Appendix C, show that even in this setting that is least advantageous to our approach, MDE-HYB yields superior assessment of experts' accuracies.

#### 6.3.2 Experiments with different learners.

An important property of MDE-HYB is that the inference element of our approach – that is, the inference of $B_j(X_i^k)$ – is model-agnostic. This property is important, given that different techniques' inductive biases are more advantageous for learning models in different domains. Consequently, our approach is applicable to assess experts' accuracies using any modeling technique that is most suitable for learning the underlying experts' domain (e.g., for determining fraud, for medical diagnoses from medical records, etc.). In general, for a given domain, the performance of alternative inductive techniques for inferring $B_j(X_i^k)$ can be evaluated over set $S$ to select the one yielding the best performance (e.g., AUC).[11] In the main Results section, we reported results using a random forest algorithm with 100 trees to induce each expert's base model, and in this section we demonstrate MDE-HYB's performance when $B_j(X_i^k)$ is inferred with additive logistic regression: LogitBoost (henceforth, LGBoost) [Friedman et al., 2000]. For scarce ground truth instances, Figure 4 shows the relative performance of MDE-HYB's performance relative to the three benchmarks (EAR, GM-GT and GM-ALL) for low- and high-quality expert workers, and also AMT real workers, respectively. In addition, Tables A4 and A5 in Appendix A.2 shows all our results, including results for settings that have abundant ground truth and statistical significance tests' results. As shown, the results indicate that MDE-HYB – when using LGBoost to infer $B_j(X_i^k)$ – exhibits the same benefits established above; namely, MDE-HYB often yields substantially better estimations, or otherwise, at least comparable estimations to ones that are possible with the benchmarks.

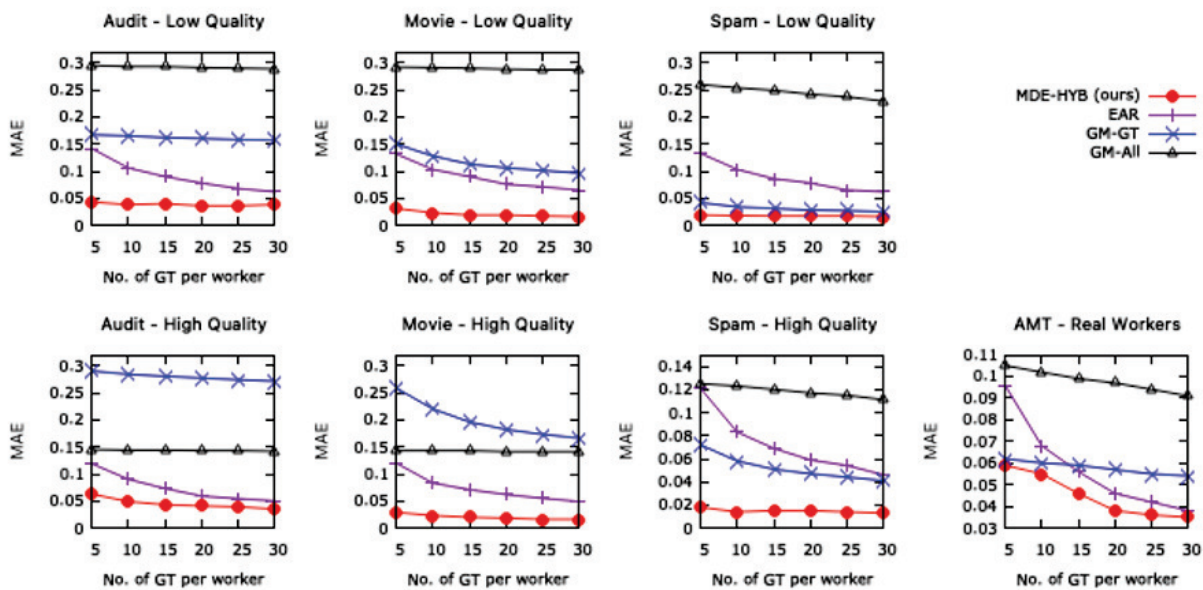

*Figure 4: MDE-HYB's Performance Relative to Benchmarks (LogitBoost used for inference by MDE-HYB, GM-GT, and GT-ALL)*

### 6.3.3 Results for Correlated Experts' Errors.

It is possible that experts' error rates are driven by properties of the decision tasks themselves, and that all experts exhibit a higher rate of error for a given set of instances. For example, such conditions correspond to settings where physicians have a higher likelihood of misdiagnosing a certain (e.g., rare) disease that all physicians have less experience diagnosing. In this section, we consider such settings in which all experts exhibit higher error rates for certain decision instances, relative to other instances. In the experiments that follow, we consider low-quality experts, all of whom exhibit a 0.2 higher likelihood of error for a given subset of decision instances. Thus, given that the experts' overall accuracies ranges between 61% and 80%, they exhibit lower accuracy for the same set of instances, ranging between 41% and 60%. In the results reported here, all experts exhibit an increased error rate for instances that have a feature value larger than the $90^{th}$ percentile of a given continuous feature.[12] The workers' overall accuracy rate remained the same as before. (Hence, workers exhibit increased accuracy for all other instances.)

---

[11]Note that this evaluation is possible because the *ranking* of different models by their performance on noisy data (the data in our setting) also correctly reflects these models' relative performances on correctly labeled data [Dekel and Shamir, 2009].

[12]The continuous features of age, star, and word frequency (word_freq_all) were used for the audit, movie, and spam datasets, respectively.

Our results are shown in Figure 5. As shown, we find that, in this setting as well, MDE-HYB either considerably outperforms the alternatives or otherwise exhibits comparable performance to them.

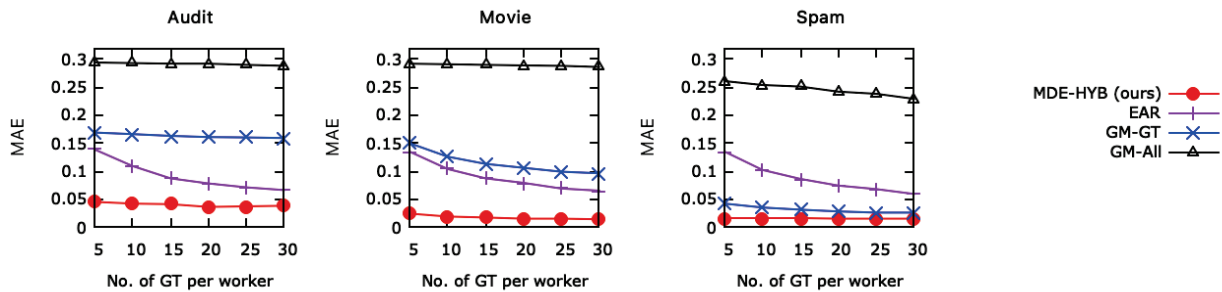

*Figure 5: MDE-HYB's Performance Relative to Benchmarks given Correlated Experts' Errors*

# 7 A Related Problem: Assessing Experts' Accuracies with an Exclusive set of Ground Truth Instances

In this section, we consider whether the approach we develop can also address a related problem: estimating experts' accuracies when scarce ground truth data are available for an exclusive set of decisions - namely, decisions that were *not* handled by any of the experts being evaluated. This problem corresponds to settings where a minimal number of ground truth decisions are provided by a third party, such as when a panel of medical experts provides a limited set of exemplary ("text book") decisions. In this setting, $GT_{exc}$ is the set of instances with ground truth labels, and $GT_{exc} \cap \{S_{W_k}\}_1^K = \emptyset$, where $\{S_{W_k}\}_1^K$ is the union of all decision instances handled by all the experts being evaluated. Although MDE-HYB was not designed for this setting, the MDE algorithm (Algorithm 1) can readily apply to this problem, with minor modifications.

In Appendix B, we outline the modifications to the MDE algorithm to address this problem, along with comprehensive evaluations of MDE's performance in this setting, using the same data domains. We find that, for this problem, MDE produces superior results over the alternatives and is the method of choice across domains and across the number of instances of ground truth.

# 8 Limitations and Future Work

We consider common experts' settings in practice, where expert workers' decision accuracy in expectation is higher than that of a random draw. However, similar to all other machine-learning-based methods, our approach may not produce an advantageous performance under unlikely pathological conditions, such as when most experts make a decision entirely at random, or when most experts are adversarial and intentionally invert their decisions.

Many machine learning methods do not lend themselves to closed-form analyses. In principle, the more complex the methods and corresponding settings are, the less likely it is that a closed form representation is possible, or that necessary abstractions can yield meaningful insights towards real world settings. Indeed, our settings involve humans decisions, and our approach involves inductions, empirical measures, and statistical procedures applied to these methods – a complexity that precludes a representation of our approach's behavior in closed form. Our results demonstrate the consistent performance of our approach: both that it is better than the alternatives considered, and the ranges of results that can be expected across domains.

It is our hope that our work will motivate future research that builds on our framework and the empirical evaluations we report here both to advance our understandings of how further improvements can be achieved and to promote the integration of these methods in practice. As the integration of machine learning in practice clearly demonstrates, such integration is essential to advance progress in practice and to identify challenges that arise in particular contexts. Our own work suggests several interesting directions for future research.

Experts' decisions are costly to acquire, and hence ground truth in these settings is inherently costly as well. Given a limited budget for ground truth acquisitions, such as via an a panel of experts, it is valuable to explore whether intelligent,information acquisition approaches for selecting instances for which to acquire ground truth information, can meaningfully improve the assessment of workers' decision accuracies. Active learning has traditionally considered selectively acquiring labels for instances with the goal of improving the generalization performance of a model induced from the acquired sample; thus, novel acquisition policies are necessary to address the goal of improving the assessment of experts decision accuracies produced by MDE-HYB or future methods developed for this problem.

Similar to most innovations, increasing transparency in experts' markets may have a variety of implications, introducing both further progress and new challenges. We previously discussed key organizational tasks that this technology can inform and advance. Here, we note possible implications that may inspire future work on related challenges. In particular, interesting and related questions that may be explored pertain to identifying particularly productive ways to bring assessments of experts' accuracies to bear in organizational settings. For example, should ongoing feedback be provided to the experts themselves to inform them about their performance, and if so, how?

In addition, depending on the context in which management would bring assessment information to bear, such as to inform managers about decisions regarding workers' retraining, it would be useful to explore if and how experts' decision performance may be affected by such practices. Studies have shown that assessments by peers or supervisors are affected by interpersonal relationships and biases [e.g. Burris, 2012]. In our approach, peers' and supervisors' evaluations are not used, and thus, any interpersonal histories and biases cannot affect the evaluation. At the same time, it would be useful to establish how different ways in which machine-learning-based assessments are brought to bear affect experts' performances. For example, given experts' knowledge that their decisions are being evaluated over time, would experts tend to improve their performances? Would workers tend to be more diligent to exhibit a consistent performance over time? Or are there conditions or kinds of tasks in which experts' performances might be undermined as a result of assessments?

## 9  Conclusions and Managerial Implications

Decision accuracy is a fundamental aspect of experts' judgment quality [Tetlock, 2017]. When ground truth is scarce, poor transparency of experts' decision accuracies undermines consumers' choices and effective management of these experts. We posit the problem and propose a framework for scalable and inexpensive data-driven assessments of experts' decision accuracies from historical data and scarce ground truth. Towards this end, we consider important settings in practice, where different experts typically decide on a mutually exclusive sets of decision instances. To our knowledge, this work is the first that addresses this challenge, proposing a general framework applicable to an arbitrary domain of expertise.

In Appendix B, we consider a related problem setting, where scarce ground truth instances are available for an exclusive, independent set of decisions that are not decided by any of the focal experts, such as decisions provided by an external and independent panel of experts. We propose a variant of our approach to address this challenge.

For both problems, we present extensive evaluations of our approach relative to alternatives, both under controlled settings using publicly available data sets, and with purposefully compiled data from human decision makers. Our empirical results show that our proposed approach is the method of choice because it yields reliable, state-of-the-art performance across a broad array of settings. Consequently, our approach can be used as a benchmark for any future works on this challenge.

The settings we consider are characteristic of key costly experts' work environments, where the ability to assess experts based on their decision accuracy provides fundamental information for management, consumers, and regulators. For management, assessment of experts' judgment accuracy is essential for a host of organizational tasks, including compensation, retention, and the optimal assignment of experts to tasks. In contrast to assessments by peer experts, our framework is scalable and inexpensive and can be readily applied to a large number of experts and decisions requiring evaluation. Also, in contrast to peer or supervisor-based assessments, which might be affected by biased human perception or prejudice, our approach is impartial and thus unaffected by such biases.

For businesses and organizations, the methodology we proposed can be readily applied to continuously assess and monitor experts' performances, and it can be used to alert experts and/or management when intervention is warranted. Similarly, for regulators, access to decision accuracy information can enable, at scale, reliable assessment of regulated services. For consumers, experts' decision accuracy is instrumental in the choice of suitable experts. Ultimately, gaining access to this information allows consumers of services to gauge the likelihood of errors and the consequential risks involved in key decisions, as well as to adequately prepare or initiate appropriate strategies to mitigate the risks, including pursuing second opinions or seeking out alternative providers altogether. Overall, given the consequences of (in)correct decisions in fields such as healthcare and security, making technology available to ascertain decision accuracy – reliably, cheaply, and at scale – is a meaningful step towards invaluable transparency.

# Appendices

## A  Additional Results for Main Ablation and Robustness Studies

In this appendix we include additional tables and results on ablation and robustness studies discussed in the main body of the paper.

### A.1  Tables for Main Ablation Studies

Our MDE-HYB approach infers $M(X_i^j)$ using an ensemble of base models. In section 6.2 we discuss two variants of inferring $M(X_i^j)$ used by our approach. One variant induces models exclusively from $GT$ (MDE-GM-GT) and another induces a global model from both $GT$ and $S$ (MDE-GM-ALL).

Table A1: MDE-HYB and Variants Performance measured by MAE for Low-quality Workers

| DATASET | GT PER WORKER | MDE-HYB | MDE GM-GT | MDE-HYB IMPROV | MDE GM-ALL | MDE-HYB IMPROV |
|---|---|---|---|---|---|---|
| Audit | 5 | **0.041** | 0.163 | 74.9%** | 0.294 | 86.0%** |
| | 10 | **0.041** | 0.159 | 74.3%** | 0.293 | 86.1%** |
| | 15 | **0.043** | 0.156 | 72.7%** | 0.292 | 85.4%** |
| | 20 | **0.036** | 0.155 | 76.8%** | 0.291 | 87.6%** |
| | 25 | **0.036** | 0.153 | 76.4%** | 0.291 | 87.6%** |
| | 30 | **0.037** | 0.151 | 75.9%** | 0.289 | 87.3%** |
| | 50 | **0.043** | 0.147 | 70.9%** | 0.285 | 85.0%** |
| | 100 | **0.034** | 0.137 | 75.0%** | 0.274 | 87.5%** |
| | 300 | **0.019** | 0.11 | 82.6%** | 0.235 | 91.8%** |
| Movie | 5 | **0.023** | 0.145 | 84.1%** | 0.293 | 92.1%** |
| | 10 | **0.017** | 0.123 | 86.0%** | 0.291 | 94.1%** |
| | 15 | **0.019** | 0.109 | 82.9%** | 0.291 | 93.6%** |
| | 20 | **0.016** | 0.099 | 84.3%** | 0.289 | 94.6%** |
| | 25 | **0.014** | 0.093 | 84.4%** | 0.288 | 95.0%** |
| | 30 | **0.015** | 0.089 | 83.1%** | 0.287 | 94.8%** |
| | 50 | **0.013** | 0.078 | 82.8%** | 0.283 | 95.3%** |
| | 100 | **0.015** | 0.065 | 76.6%** | 0.27 | 94.4%** |
| | 300 | **0.011** | 0.045 | 76.0%** | 0.223 | 95.1%** |
| Spam | 5 | **0.016** | 0.034 | 51.2%** | 0.273 | 94.0%** |
| | 10 | **0.015** | 0.029 | 47.3%** | 0.267 | 94.4%** |
| | 15 | **0.015** | 0.025 | 38.3%** | 0.262 | 94.1%** |
| | 20 | **0.015** | 0.023 | 34.1%** | 0.251 | 94.0%** |
| | 25 | **0.015** | 0.022 | 29.6%** | 0.245 | 93.8%** |
| | 30 | **0.015** | 0.021 | 26.4%** | 0.238 | 93.6%** |
| | 50 | **0.014** | 0.017 | 15.7%** | 0.214 | 93.4%** |
| | 100 | 0.014 | **0.011** | -25.8%†† | 0.149 | 90.3%** |

Experts' accuracy estimation errors. Values show Mean Absolute Error (MAE). MDE-HYB IMPROV shows the improvement of MDE-HYB over a variant. MDE-HYB yields substantially better and otherwise comparable estimations of workers accuracies. ** MDE-HYB is statistically significantly better ($p < 0.05$), *: ($p < 0.1$). ††: the other method is significantly better than MDE-HYB ($p < 0.05$). †: ($p < 0.1$).

*Table A2: MDE-HYB and Variants Performance measured by MAE for High-quality Workers*

| DATASET | GT PER WORKER | MDE-HYB | MDE GM-GT | MDE-HYB IMPROV | MDE GM-ALL | MDE-HYB IMPROV |
|---|---|---|---|---|---|---|
| Audit | 5 | **0.062** | 0.283 | 77.9%** | 0.144 | 56.5%** |
|  | 10 | **0.052** | 0.276 | 81.3%** | 0.143 | 64.1%** |
|  | 15 | **0.045** | 0.272 | 83.4%** | 0.144 | 68.6%** |
|  | 20 | **0.042** | 0.268 | 84.3%** | 0.143 | 70.6%** |
|  | 25 | **0.038** | 0.265 | 85.7%** | 0.143 | 73.5%** |
|  | 30 | **0.035** | 0.262 | 86.7%** | 0.142 | 75.4%** |
|  | 50 | **0.036** | 0.254 | 85.7%** | 0.14 | 74.1%** |
|  | 100 | **0.026** | 0.238 | 88.9%** | 0.135 | 80.5%** |
|  | 300 | **0.014** | 0.192 | 92.5%** | 0.111 | 87.0%** |
| Movie | 5 | **0.029** | 0.252 | 88.3%** | 0.143 | 79.5%** |
|  | 10 | **0.022** | 0.211 | 89.4%** | 0.144 | 84.5%** |
|  | 15 | **0.019** | 0.186 | 89.6%** | 0.143 | 86.5%** |
|  | 20 | **0.017** | 0.172 | 89.8%** | 0.142 | 87.7%** |
|  | 25 | **0.016** | 0.162 | 90.0%** | 0.142 | 88.5%** |
|  | 30 | **0.016** | 0.154 | 89.8%** | 0.142 | 88.9%** |
|  | 50 | **0.014** | 0.134 | 89.5%** | 0.139 | 89.8%** |
|  | 100 | **0.015** | 0.113 | 87.0%** | 0.133 | 88.9%** |
|  | 300 | **0.009** | 0.079 | 88.1%** | 0.103 | 90.9%** |
| Spam | 5 | **0.017** | 0.058 | 69.6%** | 0.131 | 86.7%** |
|  | 10 | **0.016** | 0.046 | 66.0%** | 0.128 | 87.8%** |
|  | 15 | **0.015** | 0.041 | 62.7%** | 0.124 | 87.7%** |
|  | 20 | **0.015** | 0.038 | 61.0%** | 0.121 | 87.9%** |
|  | 25 | **0.015** | 0.035 | 57.9%** | 0.117 | 87.3%** |
|  | 30 | **0.015** | 0.033 | 55.7%** | 0.116 | 87.4%** |
|  | 50 | **0.015** | 0.026 | 43.8%** | 0.102 | 85.7%** |
|  | 100 | **0.012** | 0.016 | 24.2%** | 0.074 | 83.1%** |

Experts' accuracy estimation errors. Values show Mean Absolute Error (MAE). MDE-HYB IMPROV shows the improvement of MDE-HYB over a variant. MDE-HYB yields substantially better and otherwise comparable estimations of workers accuracies. ** MDE-HYB is statistically significantly better ($p < 0.05$), *: ($p < 0.1$).

*Table A3: MDE-HYB and Variants Performance measured by MAE for AMT Real Workers*

| GT PER WORKER | MDE-HYB | MDE GM-GT | MDE-HYB IMPROV | MDE GM-ALL | MDE-HYB IMPROV |
|---|---|---|---|---|---|
| 5 | 0.060 | **0.058** | -3.8% | 0.107 | 43.9%** |
| 10 | **0.053** | 0.054 | 2.0% | 0.105 | 49.2%** |
| 15 | **0.046** | 0.052 | 11.5% | 0.102 | 55.3%** |
| 20 | **0.038** | 0.049 | 22.6%* | 0.1 | 61.9%** |
| 25 | **0.038** | 0.047 | 19.5%* | 0.097 | 61.2%** |
| 30 | **0.035** | 0.045 | 21.8%* | 0.095 | 62.8%** |
| 50 | **0.027** | 0.038 | 27.5%** | 0.084 | 67.6%** |
| 100 | **0.015** | 0.023 | 35.7%** | 0.032 | 52.9%** |

Experts' accuracy estimation errors. Values show Mean Absolute Error (MAE). MDE-HYB IMPROV shows the improvement of MDE-HYB over a variant. MDE-HYB yields substantially better and otherwise comparable estimations of workers accuracies. ** MDE-HYB is statistically significantly better ($p < 0.05$), *: ($p < 0.1$).

Tables A1 to A3 demonstrate the performance of MDE-HYB and the two variants. The settings include not only scarce but also abundant ground truth scenarios. Tables A1 and A2 present the results from the three semi-synthetic datasets and table A3 from the purposely compiled decision dataset (using AMT human workers). With both scarce and abundant amount of *GT*, the MDE-HYB is significantly better than these two variants of our approach in almost all cases. Generally, MDE-HYB exhibit robustness no matter how many *GT* is available across different data domains.

## A.2 Additional Results for MDE-HYB with a Different Learner

Section 6.3.2 in the main body of the paper discusses inferring base models $B_j(X_i^k)$ using a different induction algorithm. In addition to the figures provided in section 6.3.2, this appendix provides detailed numerical results for these analyses. Specifically, Tables A4 to A6 shows results of MDE-HYB using LGBoost to infer base models $B_j(X_i^k)$ including for settings with abundant ground truth.

Importantly, Tables 2 to 4 (in the main body of the paper) and tables A4 to A6 (reported here) include two sets of results using exactly the same MDE-HYB technique, except that Tables 2 to 4 reports performance using Random Forest and tables A4 to A6 reports using LGBoost to induce $B_j(X_i^k)$. The performances are considerably comparable, demonstrating that MDE-HYB is robust to the algorithm used for inducing the base models $B_j(X_i^k)$, and that in both cases MDE-HYB can achieve considerably better or at least comparable performance compared to the alternatives.

Table A4: MDE-HYB's Performance Relative to Benchmarks (LogitBoost used for inference by MDE-HYB, GM-GT, and GT-ALL) for Low-quality Workers

| DATASET | GT PER WORKER | MDE-HYB (ours) | EAR | MDE-HYB IMPROV | GM-GT | MDE-HYB IMPROV | GM-ALL | MDE-HYB IMPROV |
|---|---|---|---|---|---|---|---|---|
| Audit | 5 | **0.043** | 0.142 | 69.5%** | 0.167 | 74.0%** | 0.295 | 85.3%** |
|  | 10 | **0.038** | 0.106 | 64.0%** | 0.165 | 76.9%** | 0.293 | 87.0%** |
|  | 15 | **0.040** | 0.090 | 55.0%** | 0.162 | 75.0%** | 0.293 | 86.2%** |
|  | 20 | **0.036** | 0.078 | 54.0%** | 0.16 | 77.5%** | 0.291 | 87.6%** |
|  | 25 | **0.036** | 0.068 | 47.4%** | 0.158 | 77.2%** | 0.29 | 87.6%** |
|  | 30 | **0.038** | 0.062 | 38.7%** | 0.157 | 75.8%** | 0.289 | 86.8%** |
|  | 50 | 0.048 | **0.047** | -2.1% | 0.153 | 68.3%** | 0.285 | 83.0%** |
|  | 100 | **0.033** | 0.034 | 2.9% | 0.145 | 77.1%** | 0.275 | 87.9%** |
|  | 300 | **0.018** | 0.019 | 4.3% | 0.121 | 84.8%** | 0.235 | 92.2%** |
| Movie | 5 | **0.031** | 0.132 | 76.3%** | 0.15 | 79.1%** | 0.292 | 89.3%** |
|  | 10 | **0.023** | 0.103 | 77.7%** | 0.128 | 82.0%** | 0.291 | 92.1%** |
|  | 15 | **0.019** | 0.09 | 79.3%** | 0.113 | 83.6%** | 0.29 | 93.6%** |
|  | 20 | **0.019** | 0.076 | 75.5%** | 0.106 | 82.5%** | 0.288 | 93.6%** |
|  | 25 | **0.017** | 0.071 | 76.4%** | 0.101 | 83.4%** | 0.287 | 94.2%** |
|  | 30 | **0.016** | 0.065 | 75.0%** | 0.096 | 83.2%** | 0.287 | 94.3%** |
|  | 50 | **0.016** | 0.050 | 68.0%** | 0.085 | 81.2%** | 0.282 | 94.3%** |
|  | 100 | **0.019** | 0.035 | 46.4%** | 0.074 | 74.6%** | 0.27 | 93.0%** |
|  | 300 | **0.012** | 0.018 | 32.5%** | 0.054 | 77.2%** | 0.224 | 94.5%** |
| Spam | 5 | **0.019** | 0.133 | 85.9%** | 0.042 | 54.8%** | 0.26 | 92.7%** |
|  | 10 | **0.018** | 0.103 | 82.8%** | 0.034 | 47.1%** | 0.254 | 93.0%** |
|  | 15 | **0.017** | 0.086 | 80.7%** | 0.031 | 45.8%** | 0.249 | 93.3%** |
|  | 20 | **0.017** | 0.078 | 78.6%** | 0.028 | 40.9%** | 0.242 | 93.1%** |
|  | 25 | **0.017** | 0.065 | 74.5%** | 0.027 | 37.5%** | 0.237 | 93.0%** |
|  | 30 | **0.016** | 0.062 | 74.5%** | 0.025 | 37.6%** | 0.229 | 93.1%** |
|  | 50 | **0.016** | 0.044 | 64.1%** | 0.022 | 27.3%** | 0.208 | 92.4%** |
|  | 100 | **0.014** | 0.027 | 47.3%** | 0.014 | 2.0% | 0.15 | 90.6%** |

Experts' accuracy estimation errors. Values show Mean Absolute Error (MAE). MDE-HYB IMPROV shows the improvement of MDE-HYB over the alternative; MDE-HYB yields substantially better and otherwise comparable estimations of workers accuracies. ** MDE-HYB is statistically significantly better ($p < 0.05$), *: ($p < 0.1$). ††: the other method is significantly better than MDE-HYB ($p < 0.05$). †: ($p < 0.1$).

*Table A5: MDE-HYB's Performance Relative to Benchmarks (LogitBoost used for inference by MDE-HYB, GM-GT, and GT-ALL) for High-quality Workers*

| DATASET | GT PER WORKER | MDE-HYB (ours) | EAR | MDE-HYB IMPROV | GM-GT | MDE-HYB IMPROV | GM-ALL | MDE-HYB IMPROV |
|---|---|---|---|---|---|---|---|---|
| Audit | 5 | **0.063** | 0.119 | 47.6%** | 0.290 | 78.4%** | 0.145 | 56.7%** |
|  | 10 | **0.049** | 0.090 | 45.4%** | 0.284 | 82.6%** | 0.144 | 65.8%** |
|  | 15 | **0.043** | 0.073 | 40.4%** | 0.280 | 84.6%** | 0.144 | 69.9%** |
|  | 20 | **0.041** | 0.059 | 31.1%** | 0.277 | 85.2%** | 0.143 | 71.5%** |
|  | 25 | **0.039** | 0.054 | 27.8%** | 0.274 | 85.8%** | 0.143 | 72.8%** |
|  | 30 | **0.035** | 0.050 | 30.8%** | 0.271 | 87.1%** | 0.142 | 75.5%** |
|  | 50 | 0.038 | **0.037** | -2.9% | 0.264 | 85.4%** | 0.140 | 72.6%** |
|  | 100 | **0.026** | 0.026 | 2.1% | 0.252 | 89.8%** | 0.135 | 80.9%** |
|  | 300 | **0.014** | 0.014 | -0.4% | 0.211 | 93.2%** | 0.116 | 87.6%** |
| Movie | 5 | **0.029** | 0.120 | 75.6%** | 0.259 | 88.7%** | 0.143 | 79.5%** |
|  | 10 | **0.023** | 0.083 | 72.4%** | 0.220 | 89.5%** | 0.143 | 83.9%** |
|  | 15 | **0.020** | 0.071 | 71.6%** | 0.196 | 89.7%** | 0.143 | 85.8%** |
|  | 20 | **0.018** | 0.062 | 70.7%** | 0.182 | 90.0%** | 0.141 | 87.1%** |
|  | 25 | **0.016** | 0.055 | 70.4%** | 0.173 | 90.5%** | 0.141 | 88.3%** |
|  | 30 | **0.016** | 0.049 | 68.1%** | 0.166 | 90.5%** | 0.141 | 88.8%** |
|  | 50 | **0.015** | 0.039 | 61.1%** | 0.147 | 89.7%** | 0.138 | 89.0%** |
|  | 100 | **0.015** | 0.026 | 42.5%** | 0.127 | 88.1%** | 0.132 | 88.5%** |
|  | 300 | **0.011** | 0.014 | 17.9%** | 0.095 | 88.3%** | 0.109 | 89.9%** |
| Spam | 5 | **0.018** | 0.121 | 85.4%** | 0.072 | 75.4%** | 0.125 | 85.8%** |
|  | 10 | **0.014** | 0.083 | 82.6%** | 0.058 | 75.3%** | 0.123 | 88.3%** |
|  | 15 | **0.015** | 0.069 | 78.5%** | 0.051 | 71.2%** | 0.120 | 87.6%** |
|  | 20 | **0.015** | 0.059 | 75.4%** | 0.047 | 69.2%** | 0.117 | 87.6%** |
|  | 25 | **0.014** | 0.054 | 73.6%** | 0.044 | 68.1%** | 0.115 | 87.7%** |
|  | 30 | **0.013** | 0.046 | 70.7%** | 0.041 | 67.7%** | 0.111 | 88.0%** |
|  | 50 | **0.014** | 0.036 | 62.2%** | 0.034 | 59.7%** | 0.101 | 86.5%** |
|  | 100 | **0.012** | 0.020 | 39.6%** | 0.022 | 44.3%** | 0.072 | 83.2%** |

Experts' accuracy estimation errors. Values show Mean Absolute Error (MAE). MDE-HYB IMPROV shows the improvement of MDE-HYB using LogitBoost over an alternative. MDE-HYB yields substantially better and otherwise comparable estimations of workers accuracies. ** MDE-HYB is statistically significantly better ($p < 0.05$), *: ($p < 0.1$). ††: the other method is significantly better than MDE-HYB ($p < 0.05$). †: ($p < 0.1$).

*Table A6: MDE-HYB's Performance Relative to Benchmarks (LogitBoost used for inference by MDE-HYB, GM-GT, and GT-ALL) for AMT Real Workers*

| GT PER | MDE-HYB (ours) | EAR | MDE-HYB IMPROV | GM-GT | MDE-HYB IMPROV | GM-ALL | MDE-HYB IMPROV |
|---|---|---|---|---|---|---|---|
| 5 | **0.059** | 0.096 | 38.0%** | 0.062 | 4.9% | 0.105 | 43.4%** |
| 10 | **0.055** | 0.068 | 20.3%** | 0.060 | 9.7% | 0.102 | 46.5%** |
| 15 | **0.046** | 0.056 | 18.0%** | 0.059 | 22.4%** | 0.099 | 54.1%** |
| 20 | **0.038** | 0.046 | 18.4%** | 0.057 | 33.6%** | 0.097 | 60.9%** |
| 25 | **0.036** | 0.042 | 13.8%* | 0.055 | 35.3%** | 0.094 | 61.9%** |
| 30 | **0.035** | 0.038 | 8.7% | 0.054 | 35.6%** | 0.091 | 62.1%** |
| 50 | **0.027** | 0.027 | 0.2% | 0.047 | 41.6%** | 0.081 | 66.4%** |
| 100 | **0.015** | 0.015 | 0.0% | 0.031 | 50.6%** | 0.054 | 71.9%** |

Experts' accuracy estimation errors. Values show Mean Absolute Error (MAE). MDE-HYB IMPROV shows the improvement of MDE-HYB using LogitBoost over an alternative. MDE-HYB yields substantially better and otherwise comparable estimations of workers accuracies. ** MDE-HYB is statistically significantly better ($p < 0.05$), *: ($p < 0.1$).

## A.3 Additional Results for Settings with Correlated Workers

Section 6.3.3 discusses setting in which worker's decision errors are correlated.

Table A7: MDE-HYB's Performance Relative to Benchmarks given Correlated Experts' Errors

| DATASET | GT PER WORKER | MDE-HYB (ours) | EAR | MDE-HYB IMPROV | GM-GT | MDE-HYB IMPROV | GM-ALL | MDE-HYB IMPROV |
|---|---|---|---|---|---|---|---|---|
| Audit | 5 | **0.045** | 0.139 | 67.5%** | 0.169 | 73.3%** | 0.294 | 84.6%** |
|  | 10 | **0.042** | 0.109 | 61.7%** | 0.166 | 74.9%** | 0.293 | 85.8%** |
|  | 15 | **0.041** | 0.087 | 53.0%** | 0.163 | 75.0%** | 0.292 | 86.0%** |
|  | 20 | **0.036** | 0.078 | 54.1%** | 0.161 | 77.6%** | 0.292 | 87.7%** |
|  | 25 | **0.037** | 0.071 | 48.8%** | 0.160 | 77.2%** | 0.29 | 87.4%** |
|  | 30 | **0.038** | 0.066 | 42.0%** | 0.159 | 76.0%** | 0.288 | 86.8%** |
|  | 50 | **0.047** | 0.051 | 8.8%* | 0.154 | 69.6%** | 0.285 | 83.6%** |
|  | 100 | **0.035** | **0.035** | 0.0% | 0.147 | 76.4%** | 0.275 | 87.4%** |
|  | 300 | **0.019** | **0.019** | 0.0% | 0.123 | 84.7%** | 0.235 | 92.0%** |
| Movie | 5 | **0.024** | 0.136 | 82.3%** | 0.150 | 83.9%** | 0.292 | 91.8%** |
|  | 10 | **0.019** | 0.104 | 82.1%** | 0.126 | 85.2%** | 0.291 | 93.6%** |
|  | 15 | **0.017** | 0.087 | 80.9%** | 0.113 | 85.4%** | 0.29 | 94.3%** |
|  | 20 | **0.015** | 0.079 | 80.7%** | 0.106 | 85.5%** | 0.289 | 94.7%** |
|  | 25 | **0.015** | 0.069 | 78.7%** | 0.099 | 85.2%** | 0.288 | 94.9%** |
|  | 30 | **0.014** | 0.065 | 77.8%** | 0.096 | 85.0%** | 0.286 | 95.0%** |
|  | 50 | **0.013** | 0.05 | 73.6%** | 0.085 | 84.6%** | 0.282 | 95.3%** |
|  | 100 | **0.015** | 0.035 | 55.6%** | 0.074 | 79.3%** | 0.27 | 94.3%** |
|  | 300 | **0.011** | 0.018 | 39.3%** | 0.055 | 80.5%** | 0.223 | 95.2%** |
| Spam | 5 | **0.016** | 0.135 | 87.9%** | 0.042 | 61.2%** | 0.261 | 93.7%** |
|  | 10 | **0.016** | 0.102 | 84.7%** | 0.035 | 55.3%** | 0.254 | 93.8%** |
|  | 15 | **0.016** | 0.086 | 81.6%** | 0.031 | 49.8%** | 0.251 | 93.7%** |
|  | 20 | **0.015** | 0.074 | 80.0%** | 0.028 | 47.5%** | 0.242 | 93.9%** |
|  | 25 | **0.015** | 0.068 | 77.6%** | 0.026 | 41.1%** | 0.238 | 93.6%** |
|  | 30 | **0.015** | 0.059 | 74.9%** | 0.026 | 43.6%** | 0.229 | 93.5%** |
|  | 50 | **0.014** | 0.045 | 67.7%** | 0.021 | 30.8%** | 0.208 | 93.1%** |
|  | 100 | 0.015 | 0.027 | 43.9%** | **0.014** | -9.0%† | 0.15 | 89.8%** |

Experts' accuracy estimation errors. Values show Mean Absolute Error (MAE). MDE-HYB IMPROV shows the improvement of MDE-HYB over the alternative; MDE-HYB yields substantially better and otherwise comparable estimations of workers accuracies. ** MDE-HYB is statistically significantly better ($p < 0.05$), *: ($p < 0.1$). ††: the other method is significantly better than MDE-HYB ($p < 0.05$). †: ($p < 0.1$).

In addition to the figures provided in section 6.3.3, table A7 in this appendix provides numerical results for both scarce and abundant ground truth. By comparing table 2 (in the main body of the paper) that reports results for settings in which workers' errors are not correlated, and table A7 (in this appendix), which reports results for settings in which workers' errors are correlated – it is evident, that in most cases, MAE levels remain the same. Furthermore, results reported in table A7 demonstrate, that also in settings in which workers' errors are correlated, MDE-HYB produces superior accuracies across the three semi-synthetic data domains.[13]

### A.4 Additional Comparison between MDE-HYB and MDE

Tables A8 and A9 show a comparison between MDE-HYB and MDE methods for low and high quality semi-synthetic workers and AMT real workers respectively.

---

[13]Note that correlated error simulation was naturally conducted only with the semi-synthetic datasets and not with the purposely compiled human-decision dataset

*Table A8: Comparison between MDE-HYB and MDE with Low and High Quality Workers*

| DATASET | GT PER WORKER | Low Quality | | | High Quality | | |
|---|---|---|---|---|---|---|---|
| | | MDE-HYB | MDE | MDE-HYB IMPROV | MDE-HYB | MDE | MDE-HYB IMPROV |
| Audit | 5 | **0.041** | **0.041** | 0.0% | **0.062** | **0.062** | 0.0% |
| | 10 | **0.041** | **0.041** | 0.0% | 0.052 | **0.051** | -0.6% |
| | 15 | **0.043** | **0.043** | 0.0% | **0.045** | 0.047 | 3.8% |
| | 20 | **0.036** | 0.037 | 4.1%* | **0.042** | 0.046 | 7.6%** |
| | 25 | **0.036** | 0.039 | 6.8%** | **0.038** | 0.042 | 10.8%** |
| | 30 | 0.037 | **0.036** | -1% | **0.035** | 0.040 | 12%** |
| | 50 | 0.043 | **0.035** | -20.9%†† | **0.036** | 0.039 | 7.5%* |
| | 100 | **0.034** | 0.035 | 2.6% | **0.026** | 0.035 | 24%** |
| | 300 | **0.019** | 0.034 | 43%** | **0.014** | 0.032 | 55.5%** |
| Movie | 5 | **0.023** | **0.023** | 0.0% | **0.029** | **0.029** | 0.0% |
| | 10 | **0.017** | **0.017** | 0.0% | **0.022** | **0.022** | 0.0% |
| | 15 | **0.019** | **0.019** | 0.0% | **0.019** | **0.019** | 0.0% |
| | 20 | **0.016** | **0.016** | 0.0% | **0.017** | **0.017** | 0.0% |
| | 25 | **0.014** | **0.014** | 0.0% | **0.016** | **0.016** | 0.0% |
| | 30 | **0.015** | **0.015** | 0.0% | **0.016** | **0.016** | 0.0% |
| | 50 | **0.013** | **0.013** | 0.0% | **0.014** | **0.014** | 2.3% |
| | 100 | 0.015 | **0.013** | -17.2%†† | 0.015 | **0.013** | -13.6%†† |
| | 300 | **0.011** | 0.012 | 8.7%** | **0.009** | 0.012 | 21.9%** |
| Spam | 5 | **0.016** | **0.016** | 0.0% | **0.017** | **0.017** | 0.0% |
| | 10 | **0.015** | **0.015** | 0.0% | **0.016** | **0.016** | 0.0% |
| | 15 | **0.015** | **0.015** | 0.0% | **0.015** | **0.015** | 0.0% |
| | 20 | **0.015** | **0.015** | 0.0% | **0.015** | **0.015** | 0.0% |
| | 25 | **0.015** | **0.015** | 0.0% | **0.015** | **0.015** | 0.0% |
| | 30 | **0.015** | **0.015** | 0.0% | **0.015** | **0.015** | 0.0% |
| | 50 | **0.014** | **0.014** | 0.0% | 0.015 | **0.014** | -3% |
| | 100 | **0.014** | **0.014** | -0.6% | **0.012** | 0.013 | 7.2%* |

Experts' accuracy estimation errors. Values show Mean Absolute Error (MAE). MDE-HYB IMPROV shows the improvement of MDE-HYB over the variant MDE. MDE-HYB yields substantially better and otherwise comparable estimations of workers accuracies. ** MDE-HYB is statistically significantly better ($p < 0.05$), *: ($p < 0.1$). ††: the MDE is significantly better than MDE-HYB ($p < 0.05$). †: ($p < 0.1$).

*Table A9: Comparison between MDE-HYB and MDE with AMT Real Workers*

| GT PER WORKER | MDE-HYB | MDE | MDE-HYB IMPROV |
|---|---|---|---|
| 5 | **0.060** | **0.060** | 0.0% |
| 10 | **0.053** | 0.060 | 12.1%** |
| 15 | **0.046** | 0.059 | 23.3%** |
| 20 | **0.038** | 0.060 | 36.2%** |
| 25 | **0.038** | 0.059 | 36.5%** |
| 30 | **0.035** | 0.059 | 40.5%** |
| 50 | **0.027** | 0.060 | 54.2%** |
| 100 | **0.015** | 0.059 | 74.5%** |

Experts' accuracy estimation errors. Values show Mean Absolute Error (MAE). MDE-HYB IMPROV shows the improvement of MDE-HYB over the variant MDE. MDE-HYB yields substantially better and otherwise comparable estimations of workers accuracies. ** MDE-HYB is statistically significantly better ($p < 0.05$).

As discussed above, MDE is one of the intermediate steps used in MDE-HYB which will then decide how much weight is given to MDE and to EAR. As observed, expectedly, across all data domains, MDE-HYB and MDE present very similar results when scarce *GT* is available. This is because, in such setting when EAR is deemed by MDE-HYB to be less reliable, MDE-HYB will automatically rely heavily on MDE. Further, as observed, when datasets are highly predictable or with relatively large number of decision instances such as SPAM or MOVIE datasets - MDE-HYB and MDE obtain very similar results. Yet, Interestingly, on the more challenging datasets such as AUDIT (charecterized by low predicatbility) or AMT (charecterized by a small number of decision instances per worker), MDE-HYB generally becomes more advantageous as $GT$ becomes abundant. Or in other words, when EAR is expected to become more accurate. These results indicate that MDE-HYB makes sensical choices when giving more weight to (or solely relying on) either MDE or EAR in various conditions.

## B  Assessing Experts' Accuracies with an Exclusive set of Ground Truth Instances

In this appendix we consider a related problem: estimating experts' accuracies when scarce ground truth is available for an exclusive set of decisions, not handled by any of the experts being evaluated. This problem corresponds to settings where scarce ground truth decisions are provided by a third party, such as by a panel of medical experts that provide a limited set of exemplary ("text book") decisions. Specifically, in this setting the only instances with ground truth labels are $GT_{exc}$. Where $GT_{exc} \cap \{S_{W_k}\}_1^K = \emptyset$, and $\{S_{W_k}\}_1^K$ include all of the decisions handled by all experts being evaluated.

To address this problem we propose using a variation of the MDE algorithm (algorithm 1). Specifically, the MDE algorithm discussed in the main body of the paper can be applied with minor modifications:

a. $GT_{exc}$, the independent set of ground truth instances, replaces $GT$ (the set of workers' decisions with ground truth labels).

b. For the purpose of improving the accuracy of the base models $\{B_j\}_1^K$– instances from $GT_{exc}$ are randomly assigned and added to the training data of each base model.

The empirical evaluations reported in this appendix are similar to the empirical evaluations conducted in the main body of the paper with minor, necessary, differences. Specifically, for evaluation using semi-synthetic data sets (Audit, Movie and Spam), in order to produce an independent set of instances, that is not evaluated by any of the workers being evaluated, a subset of instances drawn uniformly at random is used to represent instances with ground truth ($GT_{exc}$), and the remaining instances are assigned to workers, as before.[14] Nevertheless, in our evaluation using the purposely compiled human workers decision dataset (decisions collected using workers on Amazon Mechanical Turk), we do not need to re-assign instances to an independent set of decisions. This is due to the fact that the original review dataset, on which this task was built upon (see appendix D) contains many instances with ground truth labels in addition to the instances randomly selected for labeling by the online workers. Therefore, to create $GT_{exc}$ we used additional instances randomly drawn from instances not labeled by the online workers. As in the main body of the paper, all evaluation procedures were repeated 50 times.

For the Audit, Movie and Spam datasets, fig. A1 shows MDE's performance for scarce number of ground truth instances, relative to alternative methods in settings with low-quality and high-quality workers. Table A10 and Table A11 show detailed results for both settings, when ground truth instances are scarce as well as abundant. The tables also detail improvements conveyed by MDE and whether such improvements are statistically significant. As shown, in this setting MDE is the method of choice: across domains, workers' qualities, and availability of ground truth, MDE exhibits state-of-the-art performance and is either superior or otherwise comparable to the alternatives. MDE assessment is significantly better than any of the alternatives across all settings (data domains, ground truth availability, and workers' performance), and at times achieves 94.9% improvement.

Recall that similar to MDE, both alternative methods, GM-GT and GM-ALL, use induced mappings to inform the expert's assessment; GM-ALL also makes use of noisy experts decisions as does MDE.[15] Our results show that MDE more effectively exploits inference, ground truth data, and expert's noisy decisions than any of the benchmarks, resulting in consistently advantageous performance across setting, unmatched by the alternatives.

---

[14] Note that, consequently, the number of instances assigned to each worker and used to build the base models in this setting is therefore smaller than in the experiments reported thus far.

[15] Note that the EAR baseline is not applicable for this problem setting as ground truth instances do not correspond to any worker's decisions

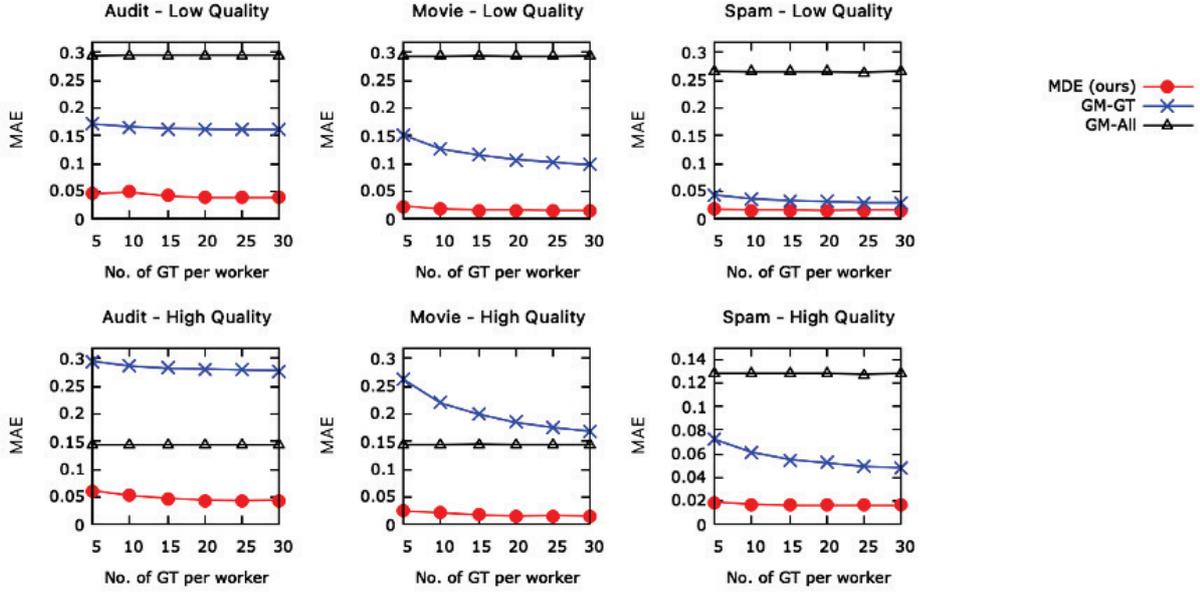

*Figure A1: MDE's Performance Relative to Benchmarks with Exclusive Set of GT*

Table A10: MDE and Benchmarks Performance measured by MAE with Exclusive Set of GT for Low-quality Workers

| DATASET | GT PER WORKER | MDE (ours) | GM-GT | MDE IMPROV | GM-ALL | MDE IMPROV |
|---|---|---|---|---|---|---|
| Audit | 5  | **0.045** | 0.171 | 73.7%** | 0.294 | 84.7%** |
|       | 10 | **0.048** | 0.166 | 71.1%** | 0.295 | 83.7%** |
|       | 15 | **0.041** | 0.163 | 74.8%** | 0.295 | 86.1%** |
|       | 20 | **0.038** | 0.162 | 76.5%** | 0.295 | 87.1%** |
|       | 25 | **0.038** | 0.161 | 76.4%** | 0.295 | 87.1%** |
|       | 30 | **0.038** | 0.161 | 76.4%** | 0.295 | 87.1%** |
| Movie | 5  | **0.023** | 0.151 | 84.8%** | 0.293 | 92.2%** |
|       | 10 | **0.018** | 0.126 | 85.7%** | 0.293 | 93.9%** |
|       | 15 | **0.016** | 0.115 | 86.1%** | 0.294 | 94.6%** |
|       | 20 | **0.016** | 0.107 | 85.0%** | 0.293 | 94.5%** |
|       | 25 | **0.015** | 0.102 | 85.3%** | 0.293 | 94.9%** |
|       | 30 | **0.015** | 0.097 | 84.5%** | 0.294 | 94.9%** |
| Spam  | 5  | **0.017** | 0.043 | 60.5%** | 0.266 | 93.6%** |
|       | 10 | **0.016** | 0.036 | 55.6%** | 0.265 | 94.0%** |
|       | 15 | **0.016** | 0.033 | 51.5%** | 0.265 | 94.0%** |
|       | 20 | **0.015** | 0.031 | 51.6%** | 0.265 | 94.3%** |
|       | 25 | **0.016** | 0.029 | 44.8%** | 0.264 | 93.9%** |
|       | 30 | **0.016** | 0.029 | 44.8%** | 0.267 | 94.0%** |

Experts' accuracy estimation errors. Values show Mean Absolute Error (MAE). MDE IMPROV shows the improvement of MDE over the alternative; MDE is the best variant. ** MDE is statistically significantly better ($p < 0.05$), *: ($p < 0.1$). ††: the other method is significantly better than MDE ($p < 0.05$). †: ($p < 0.1$).

*Table A11: MDE and Benchmarks Performance measured by MAE with Exclusive Set of GT for High-quality Workers*

| DATASET | GT PER WORKER | MDE (ours) | GM-GT | MDE IMPROV | GM-ALL | MDE IMPROV |
|---|---|---|---|---|---|---|
| Audit | 5 | **0.061** | 0.295 | 79.3%** | 0.145 | 57.9%** |
|  | 10 | **0.052** | 0.286 | 81.8%** | 0.145 | 64.1%** |
|  | 15 | **0.047** | 0.282 | 83.3%** | 0.145 | 67.6%** |
|  | 20 | **0.044** | 0.281 | 84.3%** | 0.145 | 69.7%** |
|  | 25 | **0.043** | 0.279 | 84.6%** | 0.145 | 70.3%** |
|  | 30 | **0.044** | 0.278 | 84.2%** | 0.145 | 69.7%** |
| Movie | 5 | **0.024** | 0.262 | 90.8%** | 0.144 | 83.3%** |
|  | 10 | **0.021** | 0.219 | 90.4%** | 0.144 | 85.4%** |
|  | 15 | **0.017** | 0.199 | 91.5%** | 0.145 | 88.3%** |
|  | 20 | **0.015** | 0.184 | 91.8%** | 0.144 | 89.6%** |
|  | 25 | **0.016** | 0.175 | 90.9%** | 0.144 | 88.9%** |
|  | 30 | **0.015** | 0.168 | 91.1%** | 0.144 | 89.6%** |
| Spam | 5 | **0.019** | 0.072 | 73.6%** | 0.128 | 85.2%** |
|  | 10 | **0.017** | 0.061 | 72.1%** | 0.128 | 86.7%** |
|  | 15 | **0.016** | 0.055 | 70.9%** | 0.128 | 87.5%** |
|  | 20 | **0.016** | 0.052 | 69.2%** | 0.128 | 87.5%** |
|  | 25 | **0.016** | 0.049 | 67.3%** | 0.127 | 87.4%** |
|  | 30 | **0.016** | 0.048 | 66.7%** | 0.128 | 87.5%** |

Experts' accuracy estimation errors. Values show Mean Absolute Error (MAE). MDE IMPROV shows the improvement of MDE over an alternative. MDE is the best approach. ** MDE is statistically significantly better ($p < 0.05$), *: ($p < 0.1$). ††: the other method is significantly better than MDE ($p < 0.05$). †: ($p < 0.1$).

### B.1 Evaluations with a Purposely Compiled Human Workers Decision Dataset

Figure A2 and table A12 shows the performance of MDE and other benchmarks for settings with a exclusive set of ground truth, $GT_{exc}$, using our purposely compiled human workers decision dataset (based on labels from AMT workers). As above, MDE shows lowest MAE no matter how many ground truth is provided, and is statistically significantly better than GM-ALL in all cases, and is at least as good as, or better than GM-GT.

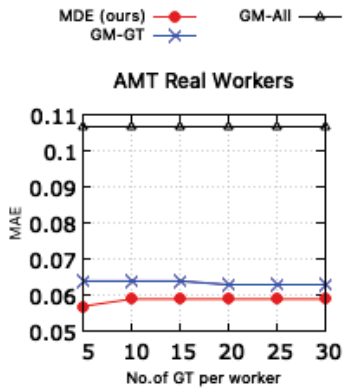

*Figure A2: Performance for Real Workers with Exclusive Set of GT*

*Table A12: Performance measured by MAE with Exclusive Set of GT for AMT Real Workers*

| GT PER WORKER | MDE (ours) | GM-GT | MDE IMPROV | GM-ALL | MDE IMPROV |
|---|---|---|---|---|---|
| 5 | **0.057** | 0.064 | 10.5% | 0.107 | 46.7%** |
| 10 | **0.059** | 0.064 | 7.5% | 0.107 | 45.0%** |
| 15 | **0.059** | 0.064 | 7.9% | 0.107 | 45.4%** |
| 20 | **0.059** | 0.063 | 6.9% | 0.107 | 44.9%** |
| 25 | **0.059** | 0.063 | 7.0% | 0.107 | 45.0%** |
| 30 | **0.059** | 0.063 | 6.7% | 0.107 | 45.0%** |
| 50 | **0.058** | 0.062 | 6.1% | 0.106 | 45.0%** |

Experts' accuracy estimation errors. Values show Mean Absolute Error (MAE). MDE IMPROV shows the improvement of MDE over an alternative. MDE is the best variant. ** MDE is statistically significantly better ($p < 0.05$), *: ($p < 0.1$). ††: the other method is significantly better than MDE ($p < 0.05$). †: ($p < 0.1$).

## C   Additional Evaluations of MDE-HYB to an Alternative

In this appendix we present the results of a comparison between experts' accuracy estimations produced by MDE-HYB and the accuracies produced as the method proposed in Tanno et al. [2019]. Tanno et al. [2019] propose a method to improve learning from noisy labelers, and that assesses labelers' errors to inform learning. Importantly, the method proposed in Tanno et al. [2019] assumes inference is done with a particular kind of models that include a cross-entropy loss function, as typically used in Neural Networks. Tanno et al. [2019] also addresses a different problem – learning such models from noisy labelers while assuming no ground truth data is available.

By contrast, MDE-HYB is model-agnostic and can be applied with any inductive technique that is most suitable for the underlying data domain and that produces class probability estimations. In addition, our problem setting focuses on inferring workers' accuracies by effectively using scarce ground truth labels. While the method proposed by Tanno et al. [2019] did not aim to address the same problem and setting we consider here, given it produces workers' accuracies, we aimed to explore the relative performance of the method proposed by Tanno et al. [2019], and using models suitable for this approch, relative to MDE-HYB's estimation of experts accuracies in settings with scarce ground truth relative. In particular, given the method proposed by Tanno et al. [2019] does not leverage ground truth, we aimed to establish whether it can yield competitive performance in setting with ground truth.

To implement approach proposed by Tanno et al. [2019], we used the TensorFlow functions and the hyperparameters provided in [Tanno et al., 2019]. To make the method proposed by Tanno et al. [2019] applicable to the data domains we use in this paper, we evaluated six different neural network architectures that can effectively apply to the different data domains in our study, corresponding to a single and two hidden layers, each with 30, 50, or 100 hidden nodes. We report the performance of the best performing neural network architecture for the method proposed by Tanno et al. [2019]. Furthermore, given our approach aims to leverage ground truth data while the method proposed in prior work does not propose how to bring such information to bear, in the experiments reported here, we considered the least favorable setting for our approach, with the minimal number of ground truth instances considered in our experiments. As before, we report average results over 50 random experiments, using different random seeds.

In Table A13, for each of the settings (data domain and workers decision quality) we report the MAE of workers' accuracy estimations produced by MDE-HYB and the approach proposed in [Tanno et al., 2019]. As shown in Table A13, across all settings, MDE-HYB's estimations of experts' accuracies are superior to those produced by the method proposed in [Tanno et al., 2019]. Given these evaluations considered the least favorable settings our approach, the results demonstrate that the method proposed in Tanno et al. [2019] does not yield competitive performance in settings with ground truth.

*Table A13: Comparison to Tanno et al.'s Alternative Baseline with GT per worker = 5*

| TASK | MDE-HYB (ours) | TANNO ET AL. Best Architecture | MDE-HYB IMPROV |
|---|---|---|---|
| Audit - Low Quality | **0.041** | 0.062 | 33.6%** |
| Audit - high Quality | **0.062** | 0.162 | 61.3%** |
| Movie - Low Quality | **0.023** | 0.202 | 88.6%** |
| Movie - high Quality | **0.029** | 0.04 | 26.4%** |
| Spam - Low Quality | **0.016** | 0.027 | 39.6%** |
| Spam - high Quality | **0.017** | 0.035 | 49.4%** |
| AMT - Real Workers | **0.06** | 0.062 | 3.87% |

Experts' accuracy estimation errors. Values show Mean Absolute Error (MAE). MDE-HYB IMPROV shows the improvement of MDE-HYB over Tanno et al.'s best architecture. MDE-HYB yields substantially better and otherwise comparable estimations of workers accuracies. ** MDE-HYB is statistically significantly better ($p < 0.05$), *: ($p < 0.1$).

# D Compilation of the Review Decision Dataset

The review texts were obtained from the well-known "Amazon fine food reviews" dataset available on Kaggle website.[16] The original dataset contains 560K reviews. From these reviews we randomly selected a subset of 10,000 reviews which meet the following criteria: (a.) Review length is between 30 to 100 words. (b.) Review was tagged with either 1 or 5 star ratings. Out of these 10,000 reviews, 8,000 reviews were used to randomly assign decision instances to forty online workers on Amazon Mechanical Turk. Each worker was assigned 200 exclusive reviews and was required to determine a binary decision of whether a review conveys a more positive or negative opinion regarding a product. Two thousand additional reviews were kept aside for the purpose of randomly selecting instances with ground truth for the purpose of evaluating the "exclusive ground truth instances" scenario (reported in Appendix B).

---
[16] https://www.kaggle.com/snap/amazon-fine-food-reviews

The star rating, originally provided by the person posting the review text, summarizes the person feeling towards this product. We followed prior work such as [Kotzias et al., 2015] which used star rating as the ground truth for review data. Nevertheless, since ground truth information is important for the correct independent evaluation of our assessment methodology, we only considered reviews that were originally tagged by the poster of the reviews with 1 or 5 stars so that we could be absolutely certain about the intent of the original poster of the review text. In our random sample of the reviews, 87.4% of the reviews were positive and 12.6% of the reviews were negative. Importantly, for making a decision, workers were only presented with the text of the reviews. We strictly kept hidden from the workers the star rating (or positive/negative labels based on the star rating). The star rating was also hidden from our method except in cases where we intentionally provide our method with (binary) scarce ground truth information. To make the task more challenging for the workers, we opened the tasks only to workers from countries which are not considered as Majority English speaking countries.[17] We also allocated only one hour to complete the task. During this time each worker had to make 200 decisions. We also verified that each worker was hired only once. For this data set, as well, we used an intentionally simple, binary bag of words representation of the reviews based on the 2,000 most popular words.

Figure A3 shows a distribution of the workers' decision accuracy as reflected by comparing their binary decision to the ground truth. As observed the majority of workers had a relatively good decision accuracy of above 85%. Though, a small number of workers had a relatively poor decision quality.

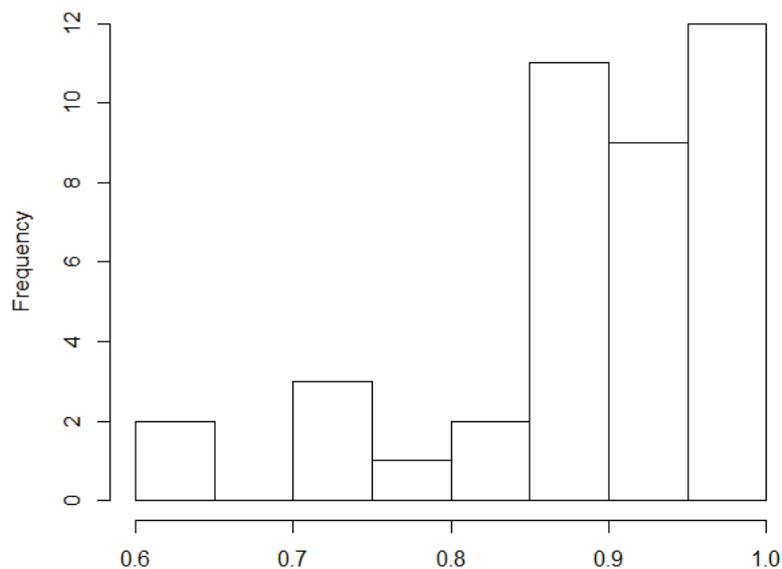

*Figure A3: Distribution of workers decision accuracy*

---

[17]https://en.wikipedia.org/wiki/English-speaking_world